\documentclass[10pt,twocolumn,letterpaper]{article}

\usepackage[pagenumbers]{cvpr}      

\usepackage{graphicx}
\usepackage{amsmath}
\usepackage{amssymb}
\usepackage{booktabs}
\usepackage{multirow}
\usepackage{gensymb}
\usepackage{makecell}
\usepackage{bm}
\usepackage{soul} 
\usepackage{color, xcolor}
\usepackage[accsupp]{axessibility}
\usepackage[pagebackref,breaklinks,colorlinks]{hyperref}
\newcommand{\ours}{Level-S$^2$fM\xspace}

\usepackage[capitalize]{cleveref}
\crefname{section}{Sec.}{Secs.}
\Crefname{section}{Section}{Sections}
\Crefname{table}{Table}{Tables}
\crefname{table}{Tab.}{Tabs.}

\def\confName{CVPR}
\def\confYear{2023}

\begin{document}

\title{\ours: Structure from Motion on Neural Level Set of Implicit Surfaces}

\author{Yuxi Xiao$^1$\qquad
Nan Xue$^{1}$\thanks{Corresponding author} \qquad
Tianfu Wu$^2$\qquad
Gui-Song Xia$^1$\\
$^1$ School of Computer Science, Wuhan University \qquad $^2$ Department of ECE, NC State University\\
{\small Project Page: \url{https://henry123-boy.github.io/level-s2fm/}}
}
\maketitle

\begin{abstract}
This paper presents a neural incremental Structure-from-Motion (SfM) approach, \ours, which estimates the camera poses and scene geometry from a set of uncalibrated images by learning coordinate MLPs for the implicit surfaces and the radiance fields from the established keypoint correspondences. 
Our novel formulation poses some new challenges due to inevitable two-view and few-view configurations in the incremental SfM pipeline, which complicates the optimization of coordinate MLPs for volumetric neural rendering with unknown camera poses. Nevertheless, we demonstrate that the strong inductive basis conveying in the 2D correspondences is promising to tackle those challenges by exploiting the relationship between the ray sampling schemes. 
Based on this, we revisit the pipeline of incremental SfM and renew the key components, including two-view geometry initialization, the camera poses registration, the 3D points triangulation, and Bundle Adjustment, with a fresh perspective based on neural implicit surfaces. 
By unifying the scene geometry in small MLP networks through coordinate MLPs,
our \ours treats the zero-level set of the implicit surface as an informative top-down regularization to manage the reconstructed 3D points, reject the outliers in correspondences via querying SDF, and refine the estimated geometries by NBA (Neural BA).
Not only does our \ours lead to promising results on camera pose estimation and scene geometry reconstruction, but it also shows a promising way for neural implicit rendering without knowing camera extrinsic beforehand.
\end{abstract}

\begin{figure}[t!]
\centering
\includegraphics[width=0.85\linewidth]{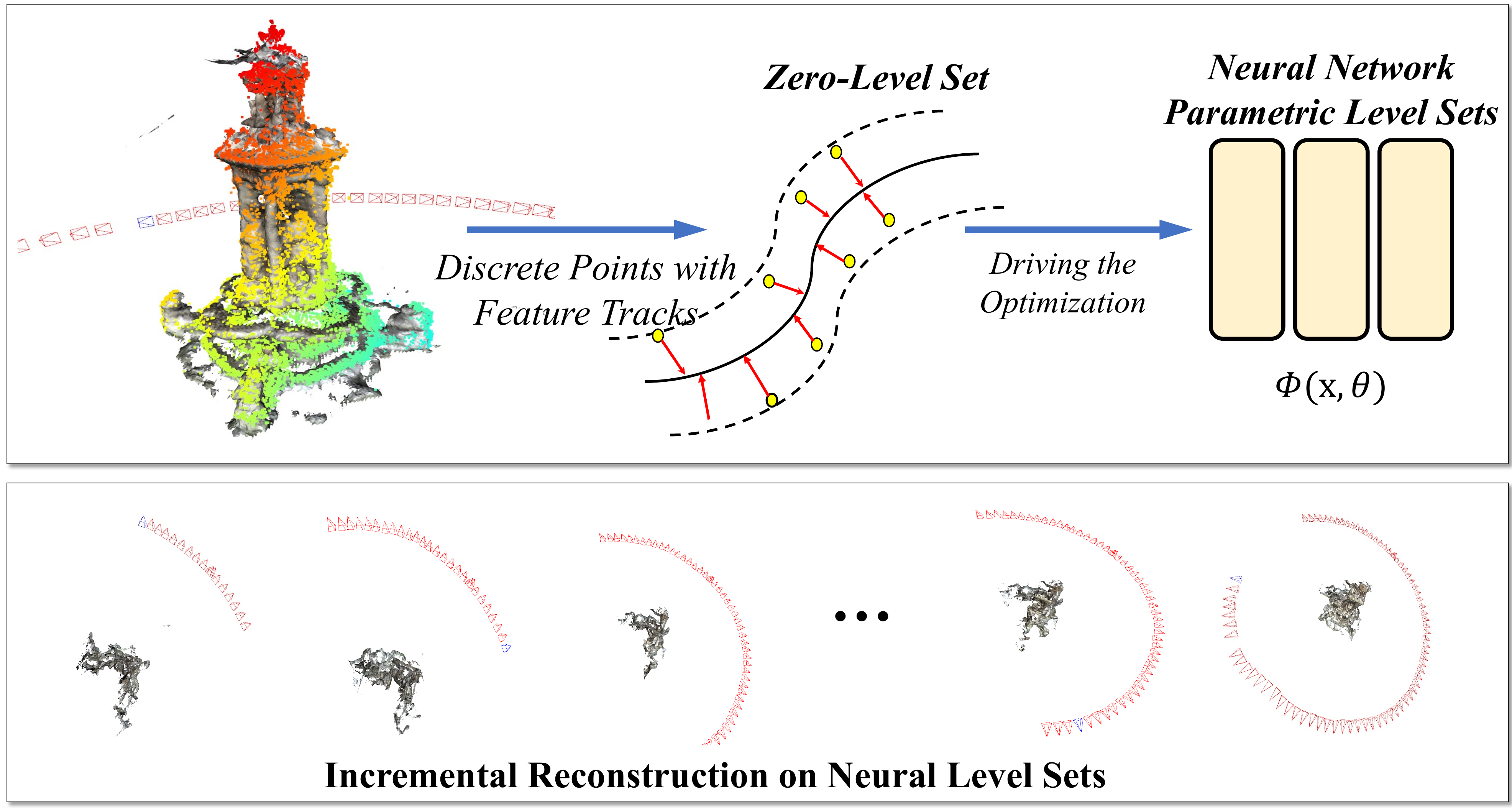}
\vspace{3mm}
\includegraphics[width=0.85\linewidth]{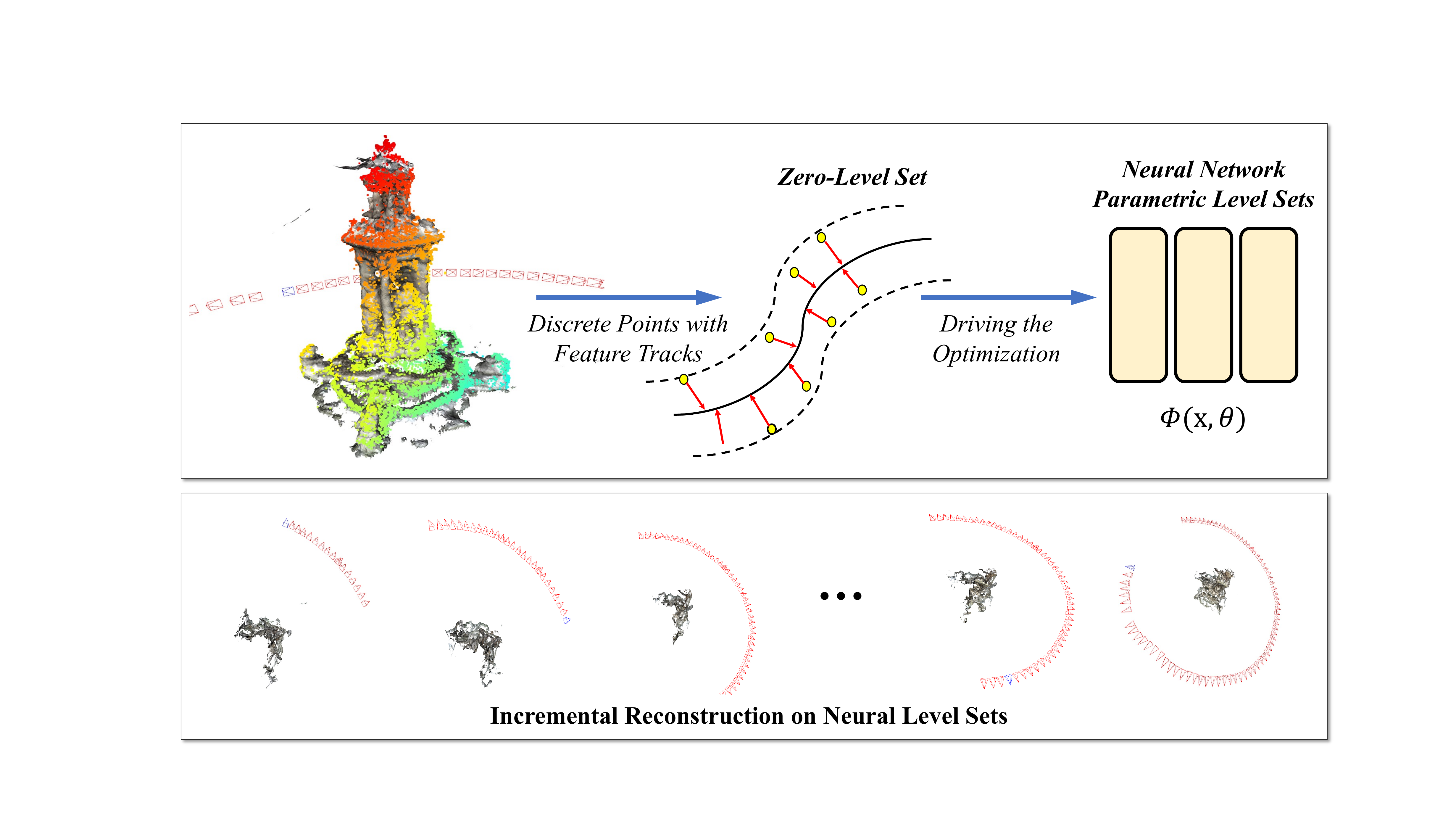}
\vspace{-4mm}
\caption{\textbf{SfM calculations on neural level sets.} We learn to do geometric calculations including Triangulation, PnP, and Bundle Adjustment above neural level sets, which easily help to reject the outliers in the matches especially in the texture repeated scenes. Also, due to the continuous surface priors of neural level sets, we achieve better pose estimation accuracy and our reconstructed points are sticking on the surface which are painted with color in the figure. While, there are a lot of outlier 3d points reconstructed by COLMAP~\cite{COLMAP} which are painted by black.}
\vspace{-6mm}
\label{fig:teaser}
\end{figure}

\vspace{-4mm}
\section{Introduction}\label{sec:Intro}
Structure-from-Motion (SfM) is a fundamental 3D vision problem that aims at reconstructing 3D scenes and estimating the camera motions from a set of uncalibrated images. As a long-standing problem, there have been a tremendous of studies that are mostly established on the keypoint correspondences across viewpoints and the theoretical findings of Multi-View Geometry (MVG)~\cite{mvg}, and have formed three representative pipelines of Incremental SfM~\cite{COLMAP}, Global SfM~\cite{globalsfm0,globalsfm1}, and Hybrid SfM~\cite{cui2017hsfm}. 

In this paper, we focus on the incremental pipeline of SfM and we will use SfM to refer to the incremental SfM.
Given an unordered image set, an SfM system initializes the computation by a pair of images that are with well-conditioned keypoint correspondences to yield an initial set of feature tracks, then incrementally adds new views one by one to estimate the camera pose from the 2D-3D point correspondences and update the feature tracks with new matches.
Because the feature tracks are generated by grouping the putative 2D correspondences across viewpoints in bottom-up manners, they would be ineffective or inaccurate to represent holistic information of scenes. Accordingly, Bundle Adjustment (BA) is necessary to jointly optimize the camera poses and 3D points in a top-down manner. The success of BA indicates that a global perspective is vital for accurate 3D reconstruction, however, their input feature tracks are the bottom-up cues without enough holistic constraints for the 3D scenes. To this end, we study to integrate the top-down information into the SfM system by proposing a novel \ours. \cref{fig:teaser} illustrates a representative case for the classic SfM systems that generate more flying 3D scene points, which can be addressed by our method.

Our \ours is inspired by the recently-emerged neural implicit surface that could manage all 3D scene points as the zero-level set of the signed distance function (SDF). Because the neural implicit surfaces can be parameterized by Multi-Layer Perceptrons (MLPs), it could be viewed as a kind of top-down information of 3D scenes and is of great potential for accurate 3D reconstruction.
However, because both the 3D scene and camera poses are to be determined, it poses a challenging problem:
\begin{quote}
   \em How can we optimize a neural SDF (or other neural fields such as NeRF) from only a set of uncalibrated images without any 3D information?
\end{quote}

Most recently, the above problem was partially answered in BARF~\cite{barf} and NeRF-\,-~\cite{nerf--} that relaxed the requirement of optimizing Neural Radiance Fields~\cite{nerf} without knowing accurate camera poses, but they can only handle the unknown pose configurations in small-scale \emph{face-forwarding} scenes. Moreover, when we confine the problem in the incremental SfM pipelines, it would be more challenging as we need to optimize the neural fields with only two overlapped images at the initialization stage. To this end, we found that the optimization of neural SDF can be accomplished by the 2D matches at the initialization stage, and facilitate the management of feature tracks by querying the 3D points and tracing the 2D keypoints in a holistic way.

As shown in \cref{fig:teaser}, we define a neural network that parameterizes an SDF as the unified representation for the underdetermined 3D scene and accomplishes the computations of PnP for camera pose intersection, the 3D points triangulation as well as the geometry refinement on the parameterized SDF. 
In the initialization stage with a pair of overlapped images, \ours uses the differentiable sphere tracing algorithm~\cite{dist_sphere} to attain the corresponding 3d points of the keypoints and calculate the reprojection error to drive the joint optimization. For the traced 3d points with small SDF values and 2D reprojection errors for its feature track, they are added into a dynamic point set and take the point set with feature tracks as the Lagrangian representation for the level sets. 
Because the pose estimation and the scene points reconstruction are sequentially estimated, the estimation errors will be accumulated. To this end, we present an NBA (\ie, Neural Bundle Adjustment) that plays a similar role as in Bundle Adjustment, but it optimizes the implicit surface and camera poses from the explicit flow of points by the energy function of the reprojection errors, which can be viewed as an evolutionary step between Lagrangian and Eulerian representations as discussed in~\cite{levelset-evolution}.

In the experiments, we evaluate our \ours on a variety of scenes from the BlendedMVS~\cite{blendedmvs}, DTU~\cite{DTU}, and ETH3D~\cite{eth3d} datasets. 
On the BlendedMVS dataset, our proposed \ours clearly outperforms the state-of-the-art COLMAP~\cite{COLMAP} by significant margins. 
On the DTU and ETH3D datasets~\cite{DTU,eth3d}, our method also obtains on-par performance with COLMAP for both camera pose estimation and dense surface reconstruction, which are all computed in one stage.

The contributions of this paper are in two folds:
\begin{itemize}
    \item We present a novel neural SfM approach \ours, which formulates to optimize the coordinate MLP networks for implicit surface and radiance field and estimate the camera poses and scene geometry. To the best of our knowledge, our \ours is the first implicit neural SfM solution on the zero-level set of surfaces.
    \item From the perspective of neural implicit fields learning, we show that the challenging problems of two-view and few-view optimization of neural implicit fields can be addressed by exploiting the inductive biases conveyed in the 2D correspondences. Besides, our method presents a promising way for neural implicit rendering without knowing camera extrinsics beforehand. 
\end{itemize}

\section{Related Works}\label{sec:method}
\subsection{Structure from Motion}
There has been a vast body of literature on Structure from Motion. Since an SfM system consists of many components, tremendous efforts have been devoted to improving the core components of SfM. 
In particular, the learning techniques were introduced in a variety of subproblems including image matching~\cite{superglue,superpoint}, feature track mining and management~\cite{TC-SfM}, two-view 3D reconstruction~\cite{two_viewsfm,DeepMLE}, relative and absolute camera pose estimation~\cite{HrubyDLP22} and Bundle Adjustment~\cite{BAnet,LSNet}. Those studies indicated that the learning paradigms are promising to improve the quality of 3D reconstruction. 
However, to the best of our knowledge, the learning paradigms are not fully equipped in SfM systems. One possible reason for such a fact is that the many learning approaches are designed in a supervised learning fashion, which remains some risks on the out-of-distribution samples. The self-supervised learning approaches~\cite{sfmlearner,monodepth} in 3D vision alleviated the requirement of data annotations, however, they have not been fully exploited in the whole pipeline of SfM.
In contrast to the aforementioned studies, in this paper, we are interested in integrating the learning ability into the SfM system without incurring any external data annotations. From the perspective of system design in SfM, we verified that the strong inductive biases conveying in the 2D correspondences are promising and meaningful to drive the learning and optimization of SfM.

\subsection{Neural Implicit Representation for 3D Scene}
Recently, the advent of neural implicit fields~\cite{deepsdf,nerf,volsdf,unisurf,neus,HFS} have greatly advanced many 3D vision problems such as novel-view synthesis~\cite{nerf,barron2021mip} and surface reconstruction~\cite{deepsdf,volsdf,unisurf,neus} by learning to optimize the coordinate MLPs from a set of posed RGB images of which the key to success is that the inductive biases of 3D are exploited by the neural networks. 
However, when the camera poses are invalid, it is hard to optimize the coordinate MLPs for neural implicit fields. To remedy this, the state-of-the-art SfM system, COLMAP~\cite{COLMAP}, is extensively used to compute the camera poses as a preprocessing step. 

To train the neural field from unknown poses directly, recently, BARF~\cite{barf} and NeRF\-\,\-~\cite{nerf--} explored to jointly optimize the camera poses and neural fields by the volumetric rendering with promising results obtained in forward-facing scenes. BARF can also work in some scenes of highly overlapped and dense image collections with the initialized poses as inputs. 
This problem was also studied in the RGB-D SLAM systems~\cite{niceslam,imap,volumetricba}, however, their works mainly rely on the known depth information and focus on the camera pose tracking by the neural implicit fields. Therefore, how to optimize implicit neural fields from only a set of uncalibrated images without any 3D information input is still a challenging and open problem. 

In this paper, we study the unknown-pose neural fields optimization and SfM together and present a unified solution that simultaneously learns the implicit surfaces and radiance fields alongside the camera pose estimation and scene reconstruction from a set of images. 

\section{Preliminaries}\label{sec:preliminary}
In this section, we introduce the preliminaries on neural implicit surface rendering and the notations in SfM, which are all extensively used in our method.
\subsection{Neural Implicit Surface Rendering}
The volumetric rendering of neural implicit surface~\cite{volsdf} aims at learning a signed distance function $d_{\Omega}: \mathbb{R}^3 \to \mathbb{R}$ by the volumetric rendering from a set of posed images and then extracting the zero-level set of $\phi$ as the reconstructed surface model of the image set. The state-of-the-art approach, VolSDF~\cite{volsdf}, integrates SDF representations with neural volume rendering via Laplacian distribution by
\begin{equation}\label{eq:sdf-distribution}
    \sigma(\mathbf{x}) = \frac{1}{\beta} \Psi_{\beta}(-d_{\Omega}(\mathbf{x})),
\end{equation}
where $\beta$ is a learnable parameter in VolSDF~\cite{volsdf}. Based on \cref{eq:sdf-distribution}, the volume rendering equation renders a ray $\mathbf{x}(t)$ emanating from a camera position $\mathbf{o}\in\mathbb{R}^3$ in unit direction $\mathbf{v}$, defined by $\mathbf{x}(t) = \mathbf{o} + t\mathbf{v}$ by
\begin{equation}\label{eq:volrend}
    I(\mathbf{o},\mathbf{v}) = \int_{0}^{\infty} L(\mathbf{x}(t), \mathbf{n}(t), \mathbf{v}) \sigma(\mathbf{x}(t)) T(t) dt,
\end{equation}
where $L(\mathbf{x},\mathbf{n},\mathbf{v})$ is the radiance field and $\mathbf{n}(t)$ is the normal direction of the point $\mathbf{x}(t)$ defined by $\mathbf{n}(t) = \nabla_{\mathbf{x}}d_{\Omega}(\mathbf{x}(t))$. 
In the learning of volume rendering, two coordinate MLP (Multi-Layer Perceptron) networks parameterize the SDF by $\mathbf{\phi}(\mathbf{x}) = (d(\mathbf{x}), \mathbf{z}(\mathbf{x})) \in \mathbb{R}^{1+256}$ and the radiance field by $L_{\psi}(\mathbf{x},\mathbf{n},\mathbf{v},\mathbf{z})\in\mathbb{R}^3$, and train them by the color loss $\mathcal{L}_{\rm RGB}(\phi,\psi,\beta)$ and the Eikonal loss $\mathcal{L}_{eik}(\phi) = \mathbb{E}_{\mathbf{z}}(\left|\nabla d(\mathbf{z})\right\|-1)$.

In this paper, we use the equations \eqref{eq:sdf-distribution} and \eqref{eq:volrend} as the basic tools for \ours. To make the optimization of SDF and radiance networks easier, we set $\beta$ as a small constant number and use the multi-resolution grid representations to avoid the potential of slow convergence and catastrophic forgetting since the scene scale is unknown and the original VolSDF~\cite{volsdf} requires to normalize the known camera poses in a certain scale.
\subsection{Ray Sampling and Sphere Tracing}
\paragraph{Iterative Ray Sampling.} In the implementation, the continuous form of \cref{eq:volrend} is approximated in 
\begin{equation}
    I(\mathbf{o},\mathbf{v}) \approx \sum_{i=1}^{m-1}\hat{\tau}_i L(\mathbf{x}(t_i), \mathbf{n}(t_i), \mathbf{v}),
\end{equation}
where $\{t_i\}_{i=1}^m$ is the discrete samples, $0=t_1<t_2<\ldots<t_m=M$, $M$ is some large constant. $\hat{\tau}_i \approx \tau(s_i)\Delta s$ is the approximated PDF multiplied by the interval length.
In VolSDF~\cite{volsdf}, $\{t_i\}_{i=1}^m$ is adaptively computed according to the opacity approximation error. Please move to \cite{volsdf} for its detail. In our method, we keep using this iterative sampling strategy when the rendering loss and the Eikonal loss is used. However, because the sampling set $\{t_i\}$ would be large, we do not use this strategy to compute the 3D points from 2D keypoints in our \ours and in turn to use the sphere tracing~\cite{1996sphere} as a faster way since our initial development of this work. 

\vspace{-4mm}
\paragraph{Sphere Tracing.} Sphere tracing is a geometric method to render the depth from a signed distance function. Different from iterative ray sampling, sphere tracing is designed to hit the surface point along the ray $\mathbf{x}(t)$ with queries as few as possible. To make it clear, we use $s_i$ to denote the ray stamp of the queried point $\mathbf{x}(s_i)$. With the queried point $\mathbf{x}(s_i)$, the next ray stamp $s_{i+1}$ is computed by $s_{i+1} = \phi(\mathbf{x}(s_i))$. In our study, we sample at most $N_s = 20$ points with the stop criterion $|\phi(s_i)|<\varepsilon$, where $\varepsilon$ is set to 0.002 in our experiment.

\paragraph{Remarks.} Although both the iterative ray sampling~\cite{volsdf} and sphere tracing~\cite{1996sphere} share the same target of computing the surface point along a ray, they have different behaviors in the neural implicit surface optimization. In detail, because VolSDF~\cite{volsdf} aims at approximating the opacity by SDF, it updates the SDF network $\phi(\mathbf{x})$ by the rendering loss $\mathcal{L}_{\rm color}$. As for sphere tracing, it is a geometric approach that only takes the SDF values into account for the computation. Such a difference is trivial to some extent, however, we found that their different focuses induce a loss function in our \ours to constraint the rendered depth values (or 3D points) for two-view initialization and 3D point triangulation.

\subsection{Notations in SfM}
\paragraph{Correspondence Search.} Given the image set $\mathcal{I} = \{I_i|i=1\ldots N_I\}$ for reconstruction, the keypoint features of the image $I_i$  computed by SIFT~\cite{SIFT} is denoted in $\mathcal{F}_i = \{(x_j, \mathbf{f}_j)\}$, where $x_j\in\mathbb{R}^2$ is the 2D coordinate and $\mathbf{f}_j \in \mathbb{R}^{128}$ is the feature descriptor of $x_j$. Based on the SIFT features, we follow the schema in COLMAP~\cite{COLMAP} to establish the feature correspondences across views, in which we first do the exhaustive matching for all possible image pairs and then use the geometric verification to filter out the non-overlapped image pairs. 
After this, the potentially overlapped image pairs are denoted in $\mathcal{C} = \{(I_a, I_b)|I_a, I_b \in \mathcal{I}\}$, and the keypoint correspondences in the pair $(I_a,I_b)$ are denoted in the set $\mathcal{M}_{ab} = \left\{\left\{(x_k,\mathbf{f}_k),(x_l',\mathbf{f}_l')\right\} | (x_k,\mathbf{f}_k \in \mathcal{F}_a, (x_l',f_l')\in\mathcal{F}_b\right\}$.
Finally, all the prepared correspondences are organized as the scene graph~\cite{scenegraph,COLMAP}, which stores images as the graph nodes and the overlapped image pairs as the graph edges.
In our \ours, we use the established correspondences to drive the learning of MLPs, estimate the camera poses, and reconstruct a sparse point set of correspondences.

\vspace{-4mm}
\paragraph{3D Scene Points and Feature Tracks.} Because SfM is designated to simultaneously estimate the scene geometry from 2D correspondences, every successfully reconstructed 3D scene point is sourced from multiple 2D keypoint observations. To facilitate the representation, we denote the expected 3D point set in $\mathcal{X} = \{\mathbf{X}_k \in \mathbb{R}^3 | k=1,\ldots,N_{3d}\}$. For each point $\mathbf{X}_k \in \mathcal{X}$, if it is reconstructed from the 2D keypoint $x_j \in \mathcal{F}_i$, we denote such a relationship in a tuple $(k,i,j)$. $\mathcal{T} = \{(k,i,j)\}$ is the set of feature tracks.

\begin{figure*}[t!]
\centering
\includegraphics[width=0.85\linewidth]{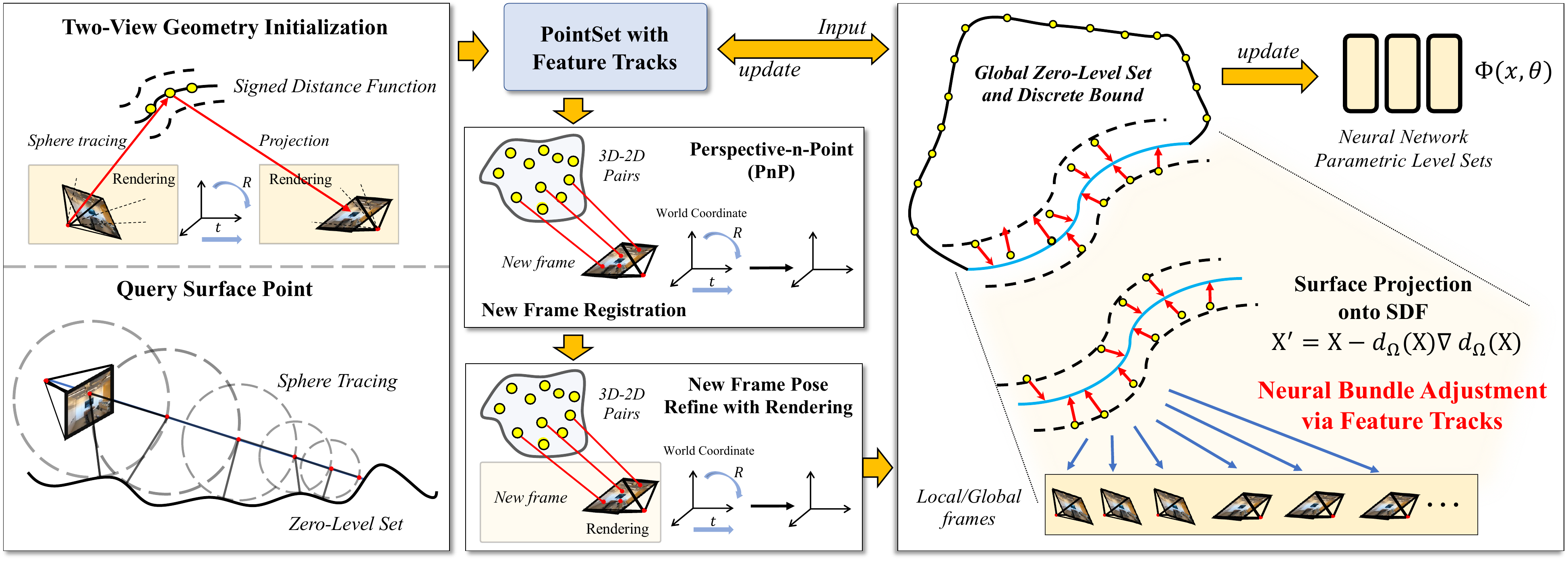}
\caption{\textbf{Overview of Level-S$^{2}$fM.}}
\label{fig:pipeline}
\end{figure*}

\section{The Proposed Level S$^2$fM}\label{sec:method}
In this section, we present the details of our \ours. As shown in Fig.~\ref{fig:pipeline}, our method consists of three classical components including 1) the two-view geometry initialization, 2) the new frame registration, and 3) the new frame pose refinement, an implicit surface and a radiance field that are parameterized by neural networks. In what follows, we will show how to solve the SfM problem by learning the implicit fields with 2D correspondences. We assume the intrinsic matrix $K$ is known and fixed.

\begin{figure}
    \centering
    \includegraphics[width=0.75\linewidth]{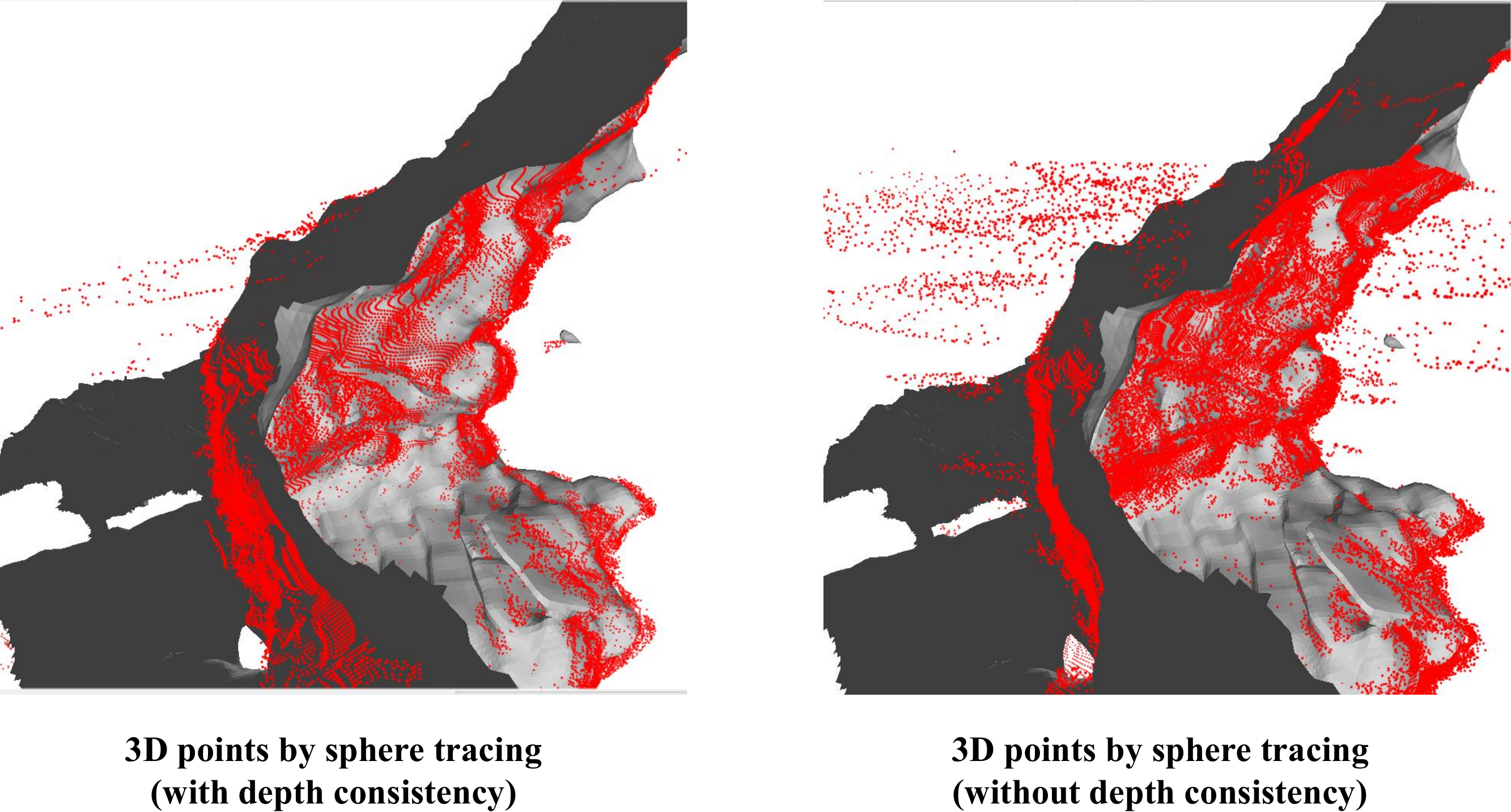}
    \caption{An illustrative comparison for the depth consistency loss $\mathcal{L}_{dc}$ in two-view initialization. As it is shown, when we remove  $\mathcal{L}_{dc}$, the 3D points traced from all putative 2D keypoints correspondences will contain more flying 3D points. }
    \vspace{-3mm}
    \label{fig:dc-compare}
\end{figure}
\subsection{Two-view Initialization}\label{sec:geoinit}
We first select two good views $\{I_{a},I_{b}\}$ for initialization from the scene graph and get their 2D matches $\mathcal{M}_{ab}=\{\{(x_k,\mathrm{f_k}),(x_l^{’},\mathrm{f}_{l}^{'})\}|(x_k,\mathrm{f_k})\in\mathcal{F}_a,(x_l^{’},\mathrm{f}_{l}^{'})\in\mathcal{F}_b\}$. Based on the 2D matches $\mathcal{M}_{ab}$, we leverage the 5-point algorithm~\cite{5points} and RANSAC to obtain the poses $P_a, P_b \in SE(3)$.

With the estimated camera poses $P_a, P_b$, it is straightforward to optimize the SDF network $\phi(\mathbf{x})$ and the radiance field network $L_{\psi}(\mathbf{x},\mathbf{n},\mathbf{v},\mathbf{z})$ defined in \cref{sec:preliminary} by minimizing the loss items $\mathcal{L}_{\rm RGB}$ and $ \mathcal{L}_{eik}$ as done in VolSDF~\cite{volsdf}. However, it should be noted that the learning of volumetric surface rendering in such a way for the \textbf{two-view inputs} would trap into the local minimal by overfitting. 
To this end, we propose to use the differentiable sphere tracking~\cite{1996sphere,dist_sphere} for the corresponding rays in image $I_a$ and $I_b$, which provides strong inductive biases for the optimization of networks.

Specifically, denoted by a pair of feature match $(x_k, x_l')$ in the image pair $(I_a,I_b)$, the sphere tracing obtains the surface point $\mathbf{X}_a^k = \mathbf{o}_a + \hat{t}_a\mathbf{d}_a$ from the SDF and $\mathbf{X}_b^l = \mathbf{o}_b + \hat{t}_b\mathbf{d}_b$, where $(\mathbf{o}_a,\mathbf{d}_a)$ is the ray of $x_k$,  $(\mathbf{o}_b,\mathbf{d}_b)$ is the ray of $x_l'$. For the computation of $\hat{t}_a$ and $\hat{t}_b$, please move to our supplementary materials. 
Ideally, the $\mathbf{X}_a^k$ and $\mathbf{X}_b^l$ should be as close as possible, therefore, we introduce a reprojection loss 
\begin{equation}
\mathcal{L}_{\text{reproj}}=\frac{1}{2V} \sum(\|\hat{x}_k-x_l'\|_{2}+\|\hat{x}_l'-x_k\|_{2}),
\label{equ:reproj}
\end{equation}
where $V$ is the number of correspondences, $\hat{x}_k = \Pi(\mathbf{X}_a^k, K, P_b)$ and $\hat{x}_l' = \Pi(\mathbf{X}_b^l, K, P_a)$ are the projected 2D coordinates of the traced 3D points by the projection $\Pi$. 

Considering the fact that the correspondences are sparse when the SDF network is not well optimized at some rays, the sparse sample points by sphere tracing on the SDF network may be either inaccurate or erroneous as shown in Fig.~\ref{fig:dc-compare}. Therefore, we use a depth consistency loss $\mathcal{L}_{dc}$ to minimize the depth estimated by the sphere tracing and the volumetric rendering by
\begin{equation}
    \mathcal{L}_{dc} = \frac{1}{B} \sum\|\hat{t}_i - \int_{0}^{\infty}T(t)\sigma(\mathbf{x}(t)) dt \|,
\end{equation}
where the rays $\mathbf{x}(t)$ are randomly sampled from the images, and those rays are also used to compute the color loss $\mathcal{L}_{\rm RGB}$. For the computation of Eikonal loss $\mathcal{L}_{eik}$, all the 3D points visited by sphere tracing and dense ray marching are used.

In summary, our two-view initialization of \ours computes the total loss $\mathcal{L}_{\rm total}^{\rm init}$ by
\begin{equation}
\mathcal{L}_{\text{total}}^{\text{init}}=\alpha_{1}\mathcal{L}_{\text{reproj}}+\alpha_{2}\mathcal{L}_{eik}+\alpha_{3}\mathcal{L}_{\rm RGB}+\alpha_{4}\mathcal{L}_{\text{dc}},
\end{equation}
where $\alpha_1$, $\alpha_2$, $\alpha_3$, and $\alpha_4$ are the hyperparameters and use ADAM optimizer to optimize the networks.

When the initialization is finished, we compute the two 3D points $\mathbf{X}$ and $\mathbf{X}'$ for each correspondence by sphere tracing for image $I_a$ and $I_b$. For an accurate correspondence, $\|\mathbf{X}-\mathbf{X}'\|$ and their SDF values should be all small enough, thus providing a good criterion to check the putative matches to initialize the 3D point set $\mathcal{X}$ and the feature track set $\mathcal{T}$ for all the verified two-view correspondences. 

\subsection{New Frame Registration} 
For every newly added frame, we will first construct the 3D-2D correspondence from the existing pointset and its feature tracks. After that, we calculate a coarse pose of the new frame with the standard PnP algorithms~\cite{epnp}, and then refine it with both the reprojection error and the rendering loss. The registration loss can be calculated as follow: 
\begin{equation}
\mathcal{L}_{\text{regist}}=\beta_{1}\mathcal{L}_{\text{reproj}}+\beta_{2}\mathcal{L}_{\rm{RGB}},
\end{equation}
where the $\beta_{1},\beta_{2}$ are two hyper-parameters, and the $\mathcal{L}_{\text{reproj}}$ here are calculated by the 3D-2D correspondences. 

In this optimization, the pose of the newly added frame, the SDF network, and the radiance field network are jointly optimized. 
While during the changes in the pose and SDF, the original location in the pointset will maybe not be the right one on the surface. For this problem, we design a Neural Bundle Adjustment (NBA) strategy to dynamically update the pointset with respect to the SDF after the points triangulation and refinement in the next section. Therefore, we leave the details of NBA in \cref{sec:nba}. 
\subsection{Points Triangulation and Refinement}
Once the pose of the newly added frame is obtained, we step into the next procedure of refining the retrieved 2D points from the point set $\mathcal{X}$ and triangulating new 2D points into 3D space to extend $\mathcal{X}$. 
This problem was formulated in the classical SfM frameworks, however, they are suffering from the following issues:
\begin{itemize}
    \item[-] \emph{The 2D Mismatches}: This issue could be alleviated by geometric verifications like RANSAC~\cite{Ransac} or better 2D keypoint matching approaches, however, when encountering the symmetry structures or repeated texture regions, those efforts are hard to work efficiently.
    \item[-] \emph{Tiny Triangulation Angle}: This issue will lead to an ill-conditioned problem for points triangulation~\cite{triangulation}. Therefore, the classical SfM approaches will directly discard those points to avoid the ill-conditioned problem configuration.  
\end{itemize}

We address those issues by proposing an SDF-based triangulation. Similar to the two-view initialization in \cref{sec:geoinit}, we compute the 3D points for all the potential 2D keypoints in the first step. Then, for the 2D keypoints that have correspondences in the current feature track set $\mathcal{T}$, we use the tracing loss $\mathcal{L}_{\rm tracing}$ 
\begin{equation}
    \mathcal{L}_{\rm tracing} = \frac{1}{V'} \sum_j \|\mathbf{X}_j^{st} - \mathbf{X}_j\|,
\end{equation}
where $\mathbf{X}_j \in \mathcal{X}$ is the retrieved 3D point of the 2D keypoint in the current frame, $V'$ is the number of retrieved 3D points. This loss function acts as the similar role of $\mathcal{L}_{dc}$ in two-view initialization. Without it, similar phenomenon like \cref{fig:dc-compare} will happen.

For the new 2D keypoints that are matched to the added images but without 3D information, both the reprojection loss similar to the two-view initialization and the tracing loss is used to yield the triangulation loss $\mathcal{L}_{\rm tri}$ by
\begin{equation}
    \mathcal{L}_{\rm tri} = \mathcal{L}_{\rm reporj}^{\rm mask} + \mathcal{L}_{\rm tracing},
\end{equation}
where $\mathcal{L}_{\rm reporj}^{\rm mask}$ only considers the 2D correspondences of which their distance between the 2D projections of the traced 3D points traced in different views are smaller than a loose threshold ($45$ pixels in our implementation).

\subsection{Neural Bundle Adjustment on Surfaces}\label{sec:nba}
Because the camera pose estimation and the points triangulation are separated, which will involve accumulative errors for  pose estimation and triangulation, as well as the implicit networks. Motivated by the Bundle Adjustment that is extensively used in classical approaches, we present a Neural Bundle Adjustment (NBA) that jointly optimizes the estimated camera points, the 3D point set, and the implicit networks as a refinement step. To avoid costly computation, our NBA step finds the closest surface points to dynamically update those variables. 

Denoted by the reconstructed 3D point set $\mathcal{X}$ and the feature track $\mathcal{T}$, the camera poses $\mathcal{P} = \{P_1, \ldots, P_K\}$ and the corresponding images $\{I_1,\ldots, I_K\}$, as well as the networks $\phi(\mathbf{x})$ and $L_{\psi}$, in each step of NBA, we update the 3D point $\mathbf{X}\in \mathcal{X}$ by 
\begin{equation}
    \mathbf{X} \leftarrow \mathbf{X} - \phi(\mathbf{X})\nabla\phi(\mathbf{X}),
    \label{equ:nba}
\end{equation}
and then compute the reprojection loss according to the feature track $\mathcal{T}$ to jointly optimize the $\phi$ the SDF network, $\mathcal{P}$ the estimated camera poses, and $\mathcal{X}$ the updated 3D point set.
For the radiance network $L_{\psi}$, the rendering loss for randomly sampled rays is computed. 

In our implementation, we leverage our NBA by three times, which we call the $1$-frame NBA, local NBA, and global NBA. 
Because the rendering loss involves more rays, we only use it for the $1$-frame NBA after the camera registration and point triangulation. In terms of local NBA, for the newly added view, only the related views with correspondences are considered. After running the $1$-frame and local NBA schemes, we globally update all reconstructed views and the point set. By leveraging the backpropagation, all the mentioned variables are updated as the refinement.

\section{Experiments}\label{sec:experiment}
\begin{table*}[]
\centering
\resizebox{.9\linewidth}{!}{
\begin{tabular}{c|c|c|c|c|c|c||c|c|c}
\toprule
\multirow{4}{*}{\makecell{Scenes}}       & \multicolumn{6}{c||}{Camera Pose Evaluation}&\multicolumn{3}{c}{Points Cloud Results Evaluation}\\\cmidrule{2-10}
                             & \multicolumn{3}{c|}{Rotation ($\degree$) $\downarrow$} & \multicolumn{3}{c||}{Translation (cm) $\downarrow$}&\multicolumn{3}{c}{\makecell{Acc(cm)$\downarrow$, Prec($\leq$3.5cm)$\uparrow$}}\\\cmidrule{2-10}
                             & COLMAP~\cite{COLMAP} & \makecell{Level-S$^2$fM\\ (full)} & \makecell{Level-S$^2$fM\\(wo/render)} & COLMAP~\cite{COLMAP} & \makecell{Level-S$^2$fM\\ (full)} & \makecell{Level-S$^2$fM\\(wo/render)}
                             & COLMAP~\cite{COLMAP} & \makecell{Level-S$^2$fM\\ (full)} & \makecell{Level-S$^2$fM\\(wo/render)}\\\midrule
LyingStatue   & 1.20      & 1.12      & 1.31      & 0.89      & 2.18      & 2.67 &\makecell{{\bf Acc}:0.83 {\bf Prec}:0.99 } & \makecell{{\bf Acc}:1.35 {\bf Prec}:0.99 }& \makecell{{\bf Acc}:1.76 {\bf Prec}:0.98 }\\
Stone       & 0.63      & 0.31      & 0.54      & 6.28      & 5.51      & 9.17 &\makecell{{\bf Acc}:5.44 {\bf Prec}:0.66 } &\makecell{{\bf Acc}:3.61 {\bf Prec}:0.68 } &\makecell{{\bf Acc}:5.04 {\bf Prec}:0.64} \\
Fountain    & 4.34      & 1.65      & 2.32      & 7.41      & 2.87      & 4.11 &\makecell{{\bf Acc}:1.91 {\bf Prec}:0.91 } & \makecell{{\bf Acc}:1.13 {\bf Prec}:0.98 }& \makecell{{\bf Acc}:1.14 {\bf Prec}:0.96 }\\
Horse       & 0.33      & 0.94      & 0.92      & 1.18      & 5.71      & 7.58 &\makecell{{\bf Acc}:3.99 {\bf Prec}:0.86 } & \makecell{{\bf Acc}:4.18 {\bf Prec}:0.71 }& \makecell{{\bf Acc}:4.18 {\bf Prec}:0.70 }\\
Statues     & 1.21      & 0.36      & 0.44      & 1.98      & 0.56      & 0.62  &\makecell{{\bf Acc}:1.31 {\bf Prec}:0.98 } & \makecell{{\bf Acc}:0.95 {\bf Prec}:0.99 }& \makecell{{\bf Acc}:1.02 {\bf Prec}:0.99 }\\\midrule
{\bf Mean}  & 1.54      &{\bf 0.86} & 1.11      & 3.54      &\bf{3.36}  & 4.83 &\makecell{{\bf Acc}:3.16 {\bf Prec}:0.85 } & \makecell{{\bf Acc}:\bf 2.25 {\bf Prec}: \bf 0.87 }& \makecell{{\bf Acc}:2.63 {\bf Prec}:0.85 }\\
\bottomrule 
\end{tabular}
}
\caption{
{\bf Quantitative results on the BlendedMVS dataset.} For our \ours, we report the results by full version and an wo/render version that removes the rendering loss during optimization.
}
\vspace{-4mm}

\label{table:blendedmvs}
\end{table*}
\subsection{Implementation Details, Datasets, and Metrics} 
\paragraph{Implementation Details.}
In our implementation, we parameterize the SDF $\phi(\mathbf{x})$ by a multi-resolution features grid and a two-layers MLP. To accelerate the computation, we follow InstantNGP~\cite{instantngp} to use a hash table~\cite{tiny-cuda-nn} for the feature grids. 
The radiance field $L_{\psi}$ is also implemented in a multi-resolution feature grid and a three-layer MLP. 
Because our end task is the geometric 3D reconstruction, we use a high-resolution multi-scale feature grid for the SDF to ensure the accuracy of scene geometry but use a low-resolution feature grid to avoid the unnecessary computation cost for the radiance field. The specifications of the network architecture are given in supplementary material due to the limited space. 
All of these above are implemented in PyTorch~\cite{pytorch}, and we used the Adam~\cite{adam} as the optimizer for the geometric calculations.  For the 2D image matching and pose graph, we keep them the same with our baseline, COLMAP~\cite{COLMAP} for fair comparisons. 

\vspace{-4mm}
\paragraph{Datasets.} Three datasets are used for our evaluation. Firstly, we use 5 representative scenes including the \emph{LyingStatue}, \emph{Stone}, \emph{Fountain}, \emph{Horse}, and \emph{Statues} from the BlendedMVS dataset~\cite{blendedmvs} in our evaluation because it provides accurate ground truth of camera poses and contains a number of challenging scenes for SfM. 
Secondly, the DTU dataset for the MVS task is also used. The five representative scenes (scans of 24, 37, 65, 110 and 114) are used in our experiments. Finally, we evaluate our proposed method on the five scenes from the challenging ETH3D~\cite{eth3d} dataset.

\vspace{-4mm}
\paragraph{Evaluation Metrics.} In our evaluation, we use the Rotation error and ATE to quantitatively benchmark the pose accuracy, which simply depicts the difference between the ground truth and the aligned pose. During our evaluation, we used the provided API of Reconstruction Align in COLMAP~\cite{COLMAP} to do that. In terms of the reconstructed scene geometry, we use accuracy (Acc) and the precision (Prec) rate to evaluate the accuracy of our recovered 3D points and Chamfer-$l1$ distance to depict the accuracy of the reconstructed surface. Detailed definitions of these evaluation metrics are given in the supplementary material.

\begin{figure}
    \centering
    \includegraphics[width=0.27\linewidth]{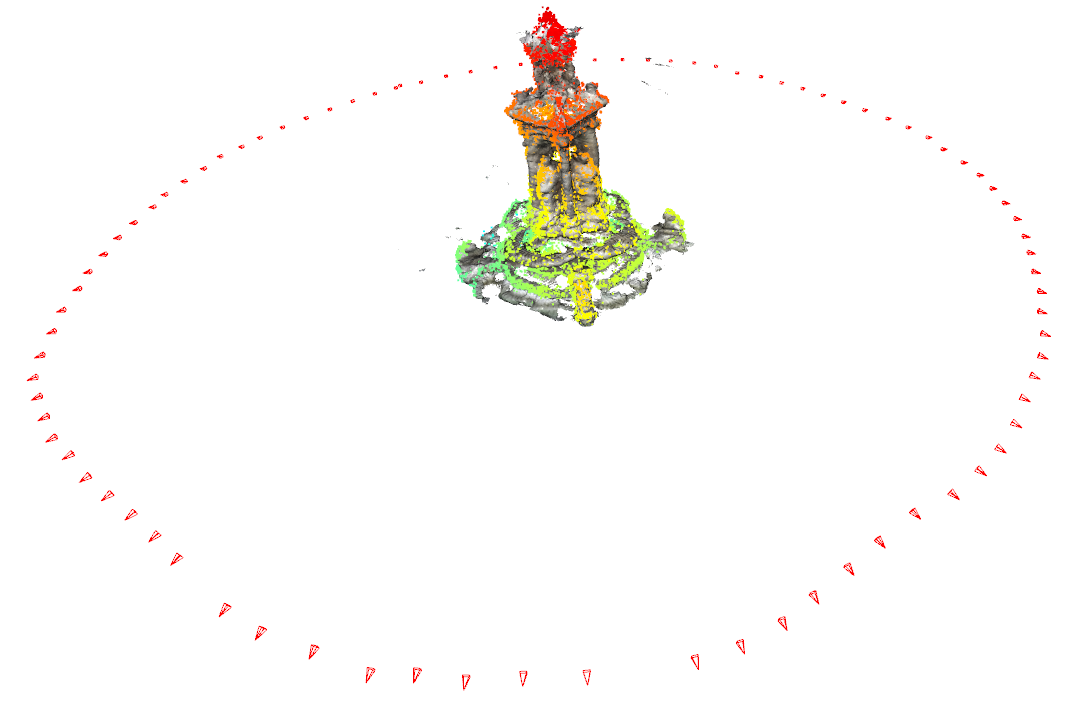}
    \includegraphics[width=0.27\linewidth]{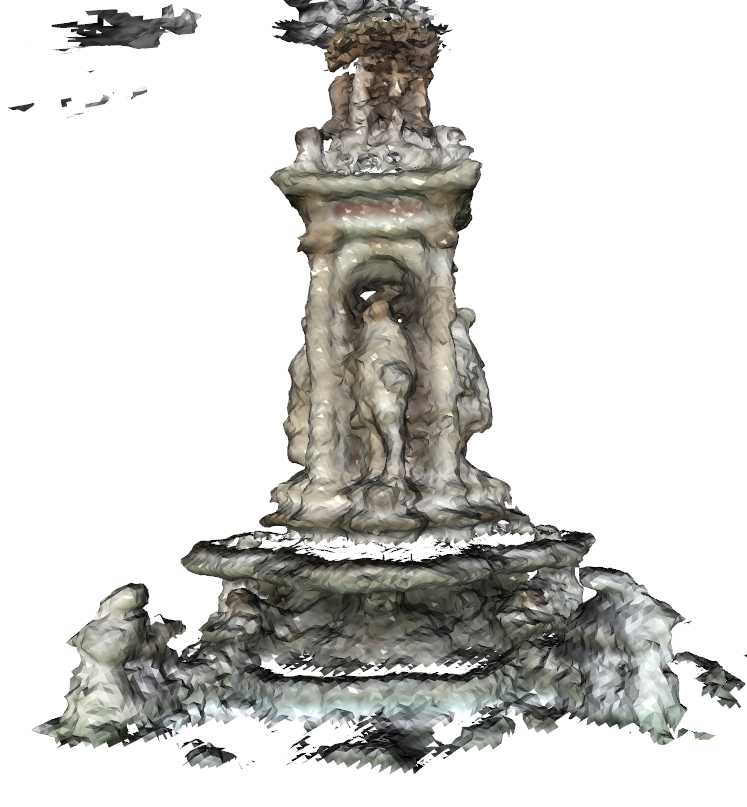}
    \includegraphics[width=0.27\linewidth]{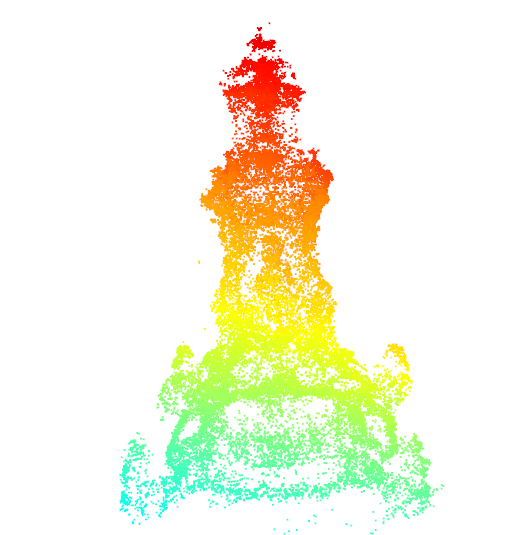}

    \includegraphics[width=0.27\linewidth]{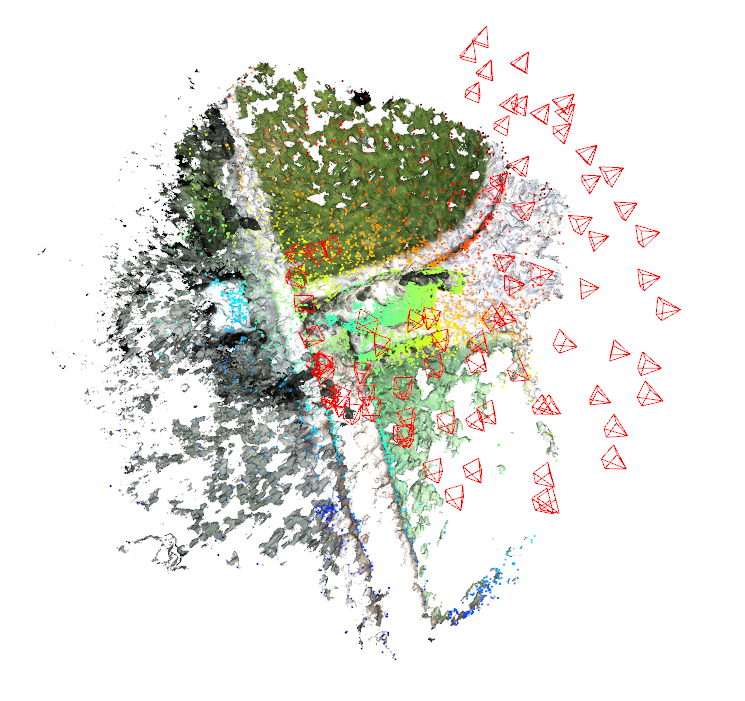}
    \includegraphics[width=0.27\linewidth]{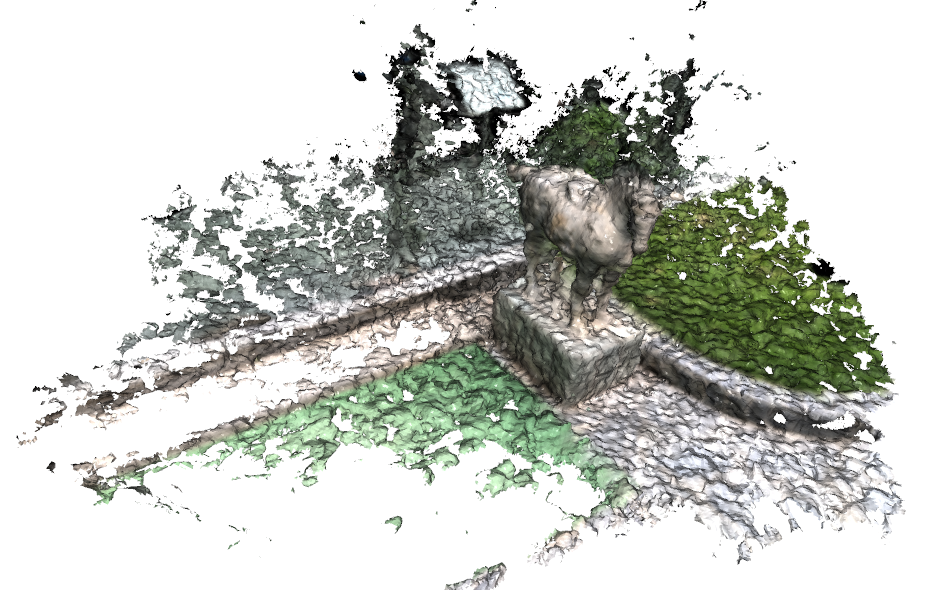}
    \includegraphics[width=0.27\linewidth]{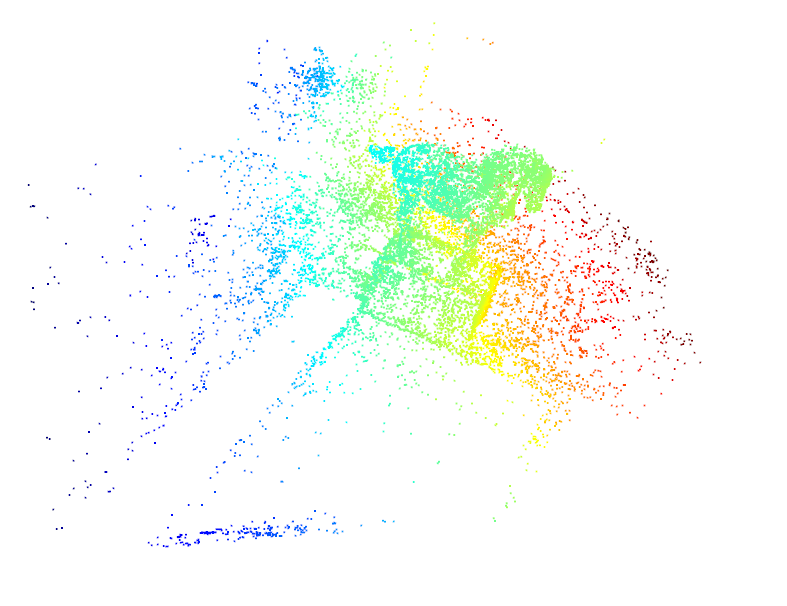}
    \caption{The reconstructed meshes, point clouds and camera poses for the \emph{Fountain} and \emph{Horse} scenes on the BlendedMVS dataset by our \ours (full). In the first column, the reconstructed scene geometry and the camera poses are shown together. For the 3D models, we show the different views of the sparse 3D points during the training and the textured meshes refused from the zero-level set surface.}
    \vspace{-5mm}
    \label{fig:bmvs-results}
\end{figure}

\subsection{Results on the BlendedMVS Dataset}
Tab.~\ref{table:blendedmvs} reports the quantitative evaluation results for the two versions of \ours and COLMAP~\cite{COLMAP}. The full version of \ours used all the mentioned components while the wo/render version removed the rendering loss for optimization. As it is reported, our \ours (full) consistently outperforms COLMAP~\cite{COLMAP} for camera pose estimation and sparse 3D point cloud reconstruction. It also reveals that rendering losses are required.

In detail, our \ours (full) averagely reduced the estimation error from 1.54$^\circ$ by COLMAP~\cite{COLMAP} to 0.86$^\circ$, \textbf{obtaining a relative improvement of 55.84\%.}  For the translation error, our \ours (full) decreases the error from 3.54~cm to 3.36~cm. For the sparse 3D point cloud reconstruction, the ACC metric is reduced from $3.16$ to $2.25$ for the full model and $2.63$ for the wo/render version.

Fig.~\ref{fig:bmvs-results} shows the reconstruction results by our method.

Apart from the direct evaluation of the SfM results on the BlendedMVS dataset, we further compare the camera poses estimation results for different methods by training the NGP~\cite{instantngp} (a fast version of NeRF~\cite{nerf}) to compare the performance of novel view synthesis in Tab.~\ref{table:tt}. As it is reported, the rendered images by our camera poses are consistently better than the ones by COLMAP poses. 

\begin{table}[]
    \resizebox{1.\linewidth}{!}{
    \begin{tabular}{c|ccccccc}
    \toprule
    \multicolumn{1}{c|}{Pose Source} &\multicolumn{1}{c}{LyingStatue}&\multicolumn{1}{c}{Stone}&\multicolumn{1}{c}{Fountain}&\multicolumn{1}{c}{Horse}&\multicolumn{1}{c}{Statues}&\multicolumn{1}{c}{Mean}
    \\
    \midrule
    \makecell{COLMAP}&{29.1}&{28.4}&{25.4}&{23.5}&{29.2}&{27.12}\\
    \makecell{\ours (Ours)}&{29.5}&{28.9}&{27.1}&{23.6}&{30.2}&{27.86}\\
    \makecell{GT}&{29.5}&{29.4}&{27.4}&{24.1}&{31.1}&{28.3}\\
    \bottomrule
    \end{tabular}}
    \vspace{-3mm}
    \caption{\textbf{Novel View Synthesis Comparison.} The PSNR is used to compare the camera poses computed by COLMAP, \ours and the GT poses on the BlendedMVS dataset. 
    }
    \vspace{-3mm}
    \label{table:tt}
\end{table}

\subsection{Results on the DTU Dataset}
\begin{table}
\centering
\resizebox{0.97\linewidth}{!}{
\begin{tabular}{c|ccc|ccc}
    \toprule
  \multirow{2}{*}{Scan}   & \multicolumn{3}{c|}{COLMAP~\cite{COLMAP}} & \multicolumn{3}{c}{\ours (Ours)} \\
         & Chamfer-$\ell_1$  & Rot. Err. & Trans. Err.  & Chamfer-$\ell_1$ & Rot. Err. & Trans. Err.\\\midrule

    24   & 2.176 & 0.38 & 2.87 & 2.442 & 0.81  & 4.60 \\
    37   & 3.837 & 0.41 & 4.86 & 3.023 & 0.31  & 4.29 \\
    65   & 4.394 & 0.45 & 4.23 & 3.190 & 0.74  & 5.81 \\
    110  & 3.389 & 0.65 & 6.36 & 5.902 & 0.82  & 6.82 \\
    114  & 3.577 & 0.35 & 3.58& 2.092  & 0.14  & 1.85 \\
    \midrule
    Mean & 3.330 & 0.448 & 4.38& 3.474  & 0.564 & 4.67 \\
    \bottomrule
\end{tabular}
}

\caption{{\bf Quantitative results on DTU dataset.} The Chamfer-$\ell_1$ distance of the dense reconstruction results and as the errors of rotation and translation for camera pose estimation, are compared for COLMAP and our \ours. The unit of Chamfer-$\ell_1$ and Translation errors are in millimetres.}
\vspace{-4mm}
\label{table:dtu}
\end{table}

We conducted the evaluation on the DTU to illustrate the promising future of our \ours to unify the pose estimation, dense reconstruction, and novel view synthesis problems in one stage. 
For the comparison to COLMAP, we use their built-in PatchMatch MVS~\cite{Patchmatch} functionality to obtain the dense surface points and then leverage its default surface reconstruction method (\ie, Poisson surface~\cite{kazhdan2006poisson}) to obtain the mesh model. For our \ours, we use the MarchingCubes~\cite{MC} to extract the mesh models from the zero-level set of the implicit surface. The quantitative evaluation results are shown in Tab.~\ref{table:dtu}. In this dataset, our \ours obtains on-par performance with COLMAP.

\subsection{Results on the ETH3D Dataset}
We test our method on a more challenging dataset, ETH3D~\cite{eth3d}, which includes both sparse view collections for multi-scale outdoor and indoor scenes. To show the influence of different keypoint detection and matching algorithms for our method, we additionally make a comparison with SuperPoint (SP)~\cite{superpoint} for detection and SuperGlue (SG)~\cite{superglue} for keypoint matching. 
As reported in Tab.~\ref{tab:eth3d}, our method achieves comparable results with COLMAP~\cite{COLMAP}. However, we observe that our method gets slightly inferior results in some large-scale outdoor scenes, because of the limited representative capability of a single network for a large-scale scene. 
\begin{table}
    \centering
    \resizebox{\linewidth}{!}{
    \begin{tabular}{c|c|c|c|c|c|c}
    \toprule
         \textbf{Scene} & \textbf{Detector \& Matcher} &courtyard&relief&door&terrace2&facade  \\
         \hline
         COLMAP~\cite{COLMAP} & \multirow{2}{*}{SIFT~\cite{SIFT}} & 0.10°/0.016m &0.10°/0.003m &0.16°/0.002m &0.14°/0.002m &0.06°/0.016m\\
         Level-S$^2$fM (Ours) & & 0.21°/0.047m &0.09°/0.003m &0.19°/0.006m &0.13°/0.003m &0.12°/0.059m \\
         \hline
         COLMAP~\cite{COLMAP} & \multirow{2}{*}{SP~\cite{superpoint}+SG~\cite{superglue}} & -- & 0.36°/0.007m & 0.37°/0.003m & 0.16°/0.003m & 0.04°/0.014m \\
         Level-S$^2$fM &  & -- & 0.42°/0.003m & 0.34°/0.007m & 0.12°/0.002m & 0.10°/0.051m \\
         \hline
    \end{tabular}
    }
    \caption{\textbf{Quantitive results} of pose estimation on the five scenes in ETH3D dataset for COLMAP~\cite{COLMAP} and our proposed \ours by using different keypoint detector and matcher.}
    \label{tab:eth3d}
\end{table}

\subsection{Ablation Study}
\begin{figure}
    \centering
    \vspace{-4mm}
    \fbox{\includegraphics[width=0.4\linewidth]{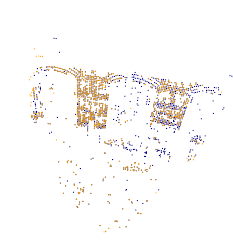}}
    \fbox{\includegraphics[width=0.4\linewidth]{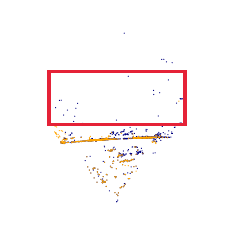}}
    \includegraphics[width=0.9\linewidth]{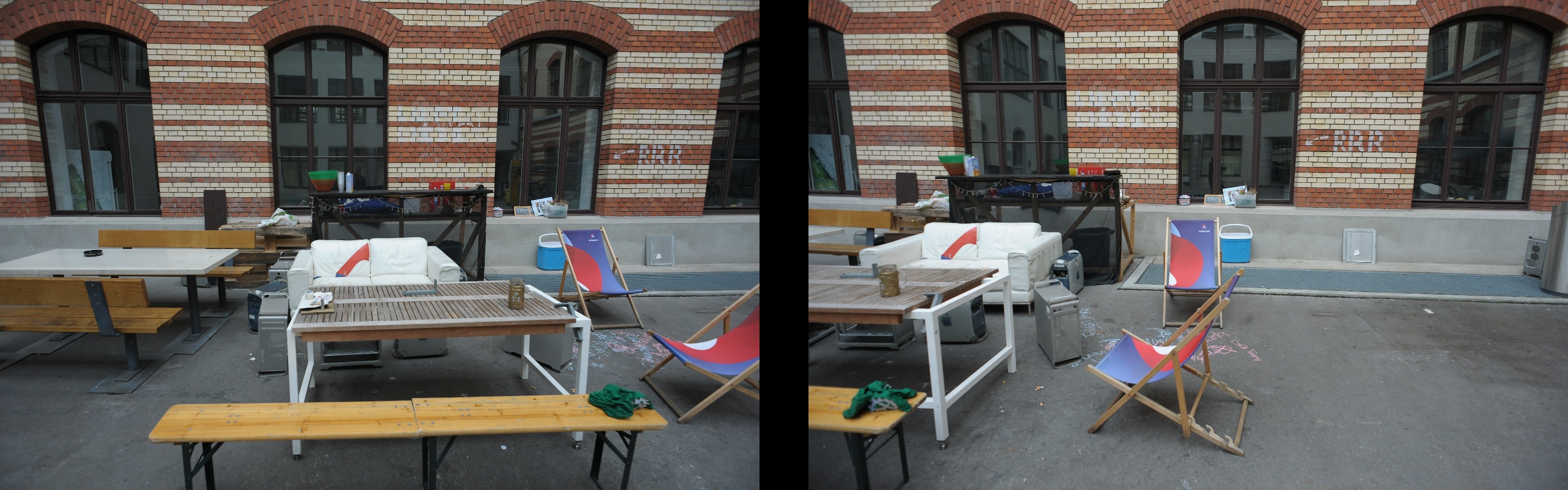}
    \resizebox{0.99\linewidth}{!}{
    \begin{tabular}{c|c|c|c}
    \toprule
    w/o SDF-Tri. & w/o NBA & w/o SDF-Tri \& w/o NBA & Full\\\hline
    1.23$^\circ$ / 0.498m & 0.46$^\circ$ / 0.113m& 3.48$^\circ$ / 0.846m&0.21$^\circ$ / 0.047m\\
    \bottomrule
    \end{tabular}
    }
    \vspace{-3mm}
    \caption{\footnotesize Two-views Triangulated Point Clouds by \textcolor{blue}{Traditional triangulation} and \textcolor{orange}{SDF-Based Triangulation}, courtyard@ETH3D.}
    \vspace{-6mm}
    \label{fig:tri_sdf}
\end{figure}

In this section, we elaborate on why and how the SDF-based Triangulation (short in {\em SDF-Tri}) and NBA work in our system. We first conduct the two-view triangulation with SDF-Tri and the traditional method respectively. Fig.~\ref{fig:tri_sdf} shows SDF-Tri can easily filter out the incorrect triangulation from wrong matches. One explanation is that the neural network distills inliers in a continuous zero-level set of surfaces (e.g., the planes in Fig.~\ref{fig:tri_sdf}). Therefore, the triangulated outliers (the blue points flying off the planes) can be easily detected and filtered by their large SDF values. Similarly, NBA also benefits from the global level sets of surfaces, which average the errors among the inliers triangulated points and differs the outliers by their large SDF values. The quantitative results for SDF-Tri and NBA are shown at the bottom of Fig.~\ref{fig:tri_sdf}. 

\subsection{Limitations of \ours }
In order to explore the clear boundary of \ours and point out the potential future development, we discuss the limitation of \ours on the most typical indoor dataset, scannet~\cite{scannet}. In the Scannet~\cite{scannet}, there are a lot of challenges including blurry images, and textureless areas. Because of the less texture, the SIFT-based keypoint correspondences may contain a large portion of outliers or insufficient matches. 
Meanwhile, the blur in images will also influence the accuracy of the 2d matches. Therefore, most SfM easily fails on this dataset. 
Our method is also limited by this because of SIFT matches. 

To make the discussion clear, we run four scenes of Scannet~\cite{scannet} that were used for NICESLAM~\cite{niceslam}. For the image sequence of each scene, the input image set for SfM is constructed by sampling for every 10 frames. 
Tab.~\ref{table:scannet} reports the pose accuracy of COLMAP and our \ours for results by adding 60 frames, 120 frames, and all the frames, which are concatenated by the ``slash". As it is reported, our method usually performs well in the first 60 frames but its pose estimation accuracy suddenly decreases when adding some new frames. 
The textureless matches cause the matches very sparse and therefore hard to give a good registration of images. Meanwhile, since the radiance field learning is also challenging in the Scannet dataset, the bad initialized pose can not be refined well by the rendering loss. All of these limitations are basically from sparse 2D image matches. Besides, we observed that the ADAM optimizer will make the optimization of camera poses and scene points unstable, which would also affect the final results.

\begin{table}
\resizebox{1.\linewidth}{!}{
\begin{tabular}{cc|c|c|c|c}
\hline
\multicolumn{2}{c}{Scene ID}&\multicolumn{1}{c}{0000}&\multicolumn{1}{c}{0059}&\multicolumn{1}{c}{0169}&\multicolumn{1}{c}{0207}\\
\hline
\multirow{2}{*}{\makecell{\textbf{iMAP~\cite{imap}}}}&{\small rot\degree $\downarrow$}&{--}&{--}&{--}&{--}
\\&{\small trans. (cm) $\downarrow$}&{197.1}&{18.9}&{96.4}&{28.7}\\
\hline
\multirow{2}{*}{\makecell{\textbf{NICESLAM~\cite{niceslam}}}}&{\small rot\degree $\downarrow$}&{--}&{--}&{--}&{--}
\\&{\small trans. (cm) $\downarrow$}&{11.3}&{12.0}&{12.0}&{12.8}\\
\hline
\multirow{2}{*}{\makecell{\textbf{COLMAP~\cite{COLMAP}}}}&{\small rot\degree $\downarrow$}&{1.47/2.71/2.71}&{1.40/2.52/2.66}&{2.64/1.84/2.52}&{failed}
\\&{\small trans. (cm) $\downarrow$}&{8.20/16.8/14.1}&{7.20/10.8/11.5}&{10.6/9.22/10.6}&{--}\\
\hline
\multirow{2}{*}{\makecell{\textbf{Level-S$^2$fM}\\full}}&{\small rot\degree $\downarrow$}&{1.44/2.23/1.98}&{1.50/2.59/3.72}&{2.94/9.09/14.3}&{failed}
\\&{\small trans. (cm) $\downarrow$}&{6.50/11.8/26.7}&{8.20/32.9/35.7}&{6.35/14.3/44.8
}&{--}\\
\bottomrule 
\end{tabular}
}
\caption{{\bf Quantitative results} of pose estimation on ScanNet~\cite{scannet}. For the COLMAP and \ours, we report their pose accuracy metrics when the 60/120/all frames are registered. For the last scene (\ie, 0207), both COLMAP and our method failed. 
}
\vspace{-3mm}
\label{table:scannet}
\end{table}

\section{Conclusion}\label{sec:conclusion}
This paper studies the longstanding problem of Structure-from-Motion by exploring and exploiting several important yet challenging issues including the two-view neural rendering in the  initialization stage and few-view neural rendering in the early camera registration stage of incremental SfM for integrating the recent advances of neural implicit field learning into an SfM pipeline. We show that although the few-view neural rendering problem is challenging enough, it can be tackled by the 2D correspondences as they convey strong inductive biases for 3D scenes. Based on this, we present the first neural SfM solution that renews several key components of two-view geometry initialization, camera pose registration, and triangulation, as well as the Bundle Adjustment problem with neural implicit fields. In the experiments, we show that \ours outperforms the traditional SfM pipeline and set a new state-of-the-art for 3D reconstruction on the BlendedMVS dataset. 
We believe that our study will encourage the 3D vision community to rethink and reformulate Structure-from-Motion with learning-based new findings. 

{\small
\paragraph{Acknowledgments.}
This work was supported by the National Nature Science Foundation of China under grants 62101390 and U22B2011. This work was also supported by Wuhan University-Huawei Geoinformatics Innovation Laboratory. Y. Xiao would love to thank Sida Peng for the helpful discussions. T. Wu was supported by NSF IIS-1909644, IIS-1822477, CMMI-2024688, and IUSE-2013451.  We are also grateful for the constructive comments by anonymous reviewers and area chairs. 
The views presented in this paper are those of the authors and should not be interpreted as representing any funding agencies.
}

\section*{Supplementary Material}
\appendix
\section{Network Architecture Details}
\begin{figure}
    \centering
    \includegraphics[width=.48\textwidth]{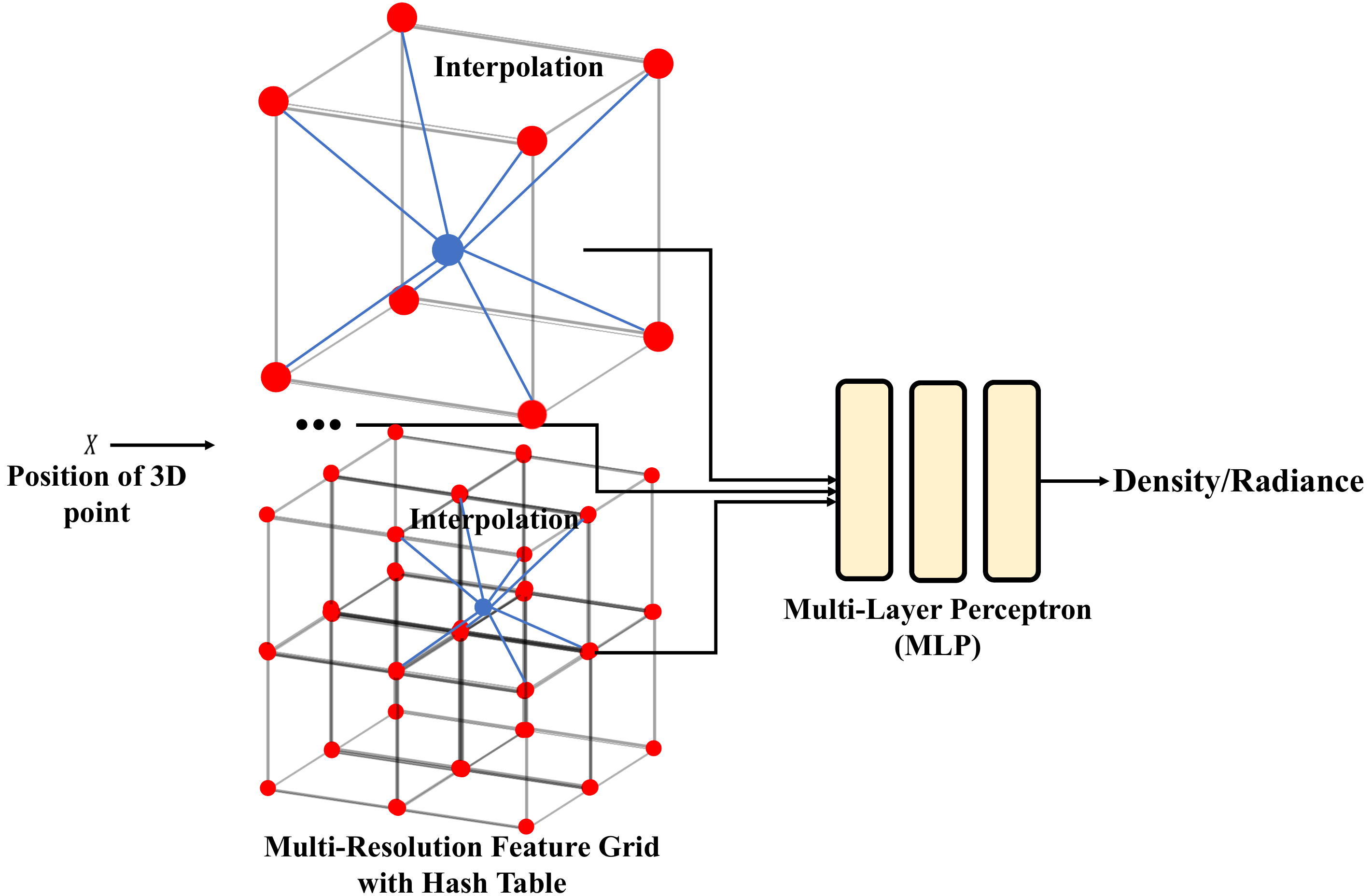}
    \caption{\textbf{The Multi-resolution Features Grid.}}
    \label{fig:arch}
\end{figure}

The architecture of our \ours is shown in Fig.~\ref{fig:arch}. We use dual fields to independently represent the radiance field and signed distance field (SDF), which have the same architecture. For each queried 3d point, we will first interpolate the feature of the queried points at multi-resolution grids, and then concatenate the multi-resolution features into the MLP to attain the density or the radiance. To accelerate the training, we adopt the multi-resolution hash table~\cite{instantngp} in our implementation. In detail, we construct multiresolution grids of $L$ levels, and the resolution of each level is:
\begin{align}
    N_l &:= \left\lfloor N_{min} \cdot b^l \right\rfloor \,,\\
    b &:= \exp\left( \frac{\ln{N_{max}} - \ln{N_{min}}}{L-1} \right) \,,\label{Eqn:PerLevelScale}
\end{align}
where $N_{min}$ and $N_{max}$ are the coarsest and finest resolutions. In the multi-resolution hash table, to obtain the feature of point $\mathbf{x}$, we first scale and round $\mathbf{x}$ at each level $l$ as $\lfloor \mathbf{x}_l\rfloor = \lfloor \mathbf{x} \cdot N_l \rfloor$, $\lceil \mathbf{x}_l\rceil = \lceil \mathbf{x} \cdot N_l \rceil$. Then we can obtain the voxel spanned by ${\lfloor \mathbf{x}_l\rfloor}$ and ${\lceil \mathbf{x}_l\rceil}$ and map each corner of the voxel to the hash table using the spatial hash function:
\begin{equation} \label{Eqn:HashFunc}
    h(\mathbf{x}) = \left(\bigoplus_{i=1}^{3} x_i \pi_i \right) \mod T \,,
\end{equation}
where $\oplus$ denotes the bit-wise XOR operation, and $\pi_i$ are unique, large prime numbers. In our implementation, $\pi_1$, $\pi_2$, $\pi_3$ and $T$ are set to $1, 2654435761, 805459861, 2^{19}$ respectively. After that, the feature vectors at each corner are interpolated at $\mathbf{x}$ by the interpolation weight $\mathbf{w}_l=\mathbf{x}_l-\lfloor \mathbf{x}_l\rfloor$. Lastly, we concatenate the feature vector of $\mathbf{x}$ at each level, as well as the encoded view direction $\mathbf{v}$ togehter, and send it into an MLP to predict the values. In our implementation, we leverage two different resolution hash tables to represent the sdf and radiance fields respectively. For sdf function, the configuration is $L=8, N_{min}=16, N_{max}=2048$ and the number of features at each level is 4. While the configuration for radiance field is $L=16, N_{min}=16, N_{max}=2048$ and the number of features at each level is 2.
 
\section{Evaluation Metrics and Details}
Because the world coordinate system varies for different SfM systems, we need to align the estimated poses to the ground truth poses first. We use the reconstruction alignment API from COLMAP~\cite{COLMAP}, which first pre-aligns the two reconstructions with their poses and refine that by aligning the sparse point clouds of them. Here, the sparse point clouds are triangulated by the 2d matches of SIFT with the fixed poses, which can be also easily implemented with the existing COLMAP API. After the alignment, the rotation error is computed as follows: 
\begin{equation}
    \theta^{\text{error}}_{i}=\text{cos}^{-1}\frac{\text{trace}(\mathbf{R}_{i}^{\text{gt}}\hat{\mathbf{R}}_{i}^{T})-1}{2}, i=\{1,...,M\},
\label{equ:rot_error}
\end{equation}
where the $M$ is the number of the cameras, and $\mathbf{R}_{i}^{\text{gt}}$, $\hat{\mathbf{R}_{i}}^{T}$ are the gt rotation matrix and aligned estimated rotation matrix respectively. We take the average error of the rotation in Equation (\ref{equ:rot_error}) as the metric for rotation. As for the evaluation of translation, we use the ATE RMSE~\cite{rgbd_eval} to depict the distance between the ground truth trajectories and the estimated, specifically followed: 
\begin{equation}
    \text{RMSE}(\hat{\mathbf{T}}_{i})=\left(\frac{1}{M}\sum^{M}_{i=0}||\text{trans}(\mathbf{T}_{i}^{-1}\hat{\mathbf{T}}_{i})||^2\right)^{\frac{1}{2}},
\end{equation}
where the $\mathbf{T}_{i},\hat{\mathbf{T}}_{i}$ are the gt and aligned transformation matrix respectively, and the $\text{trans}$ means to take the translation part of the transformation matrix. 

For the evaluation of reconstruction results, the definitions of metrics are shown in Table.~\ref{tab:my_label}. We use these two metrics to evaluate the accuracy of the reconstructed point cloud.
\begin{table}
    \centering
    \begin{tabular}{cc}
    \toprule
    Metric & Definition \\
    \hline
    Acc& $\text{mean}_{p\in P}(\text{min}_{p^{*}\in P^{*}||p-p^{*}||})$\\
    Prec&$\text{mean}_{p\in P}(\text{min}_{p^{*}\in P^{*}||p-p^{*}||\leq.035})$\\
    \bottomrule
    \end{tabular}
    \caption{Caption}
    \label{tab:my_label}
\end{table}

\section{Two View Initialization}
\begin{figure}
    \centering
    \includegraphics[width=0.8\linewidth]{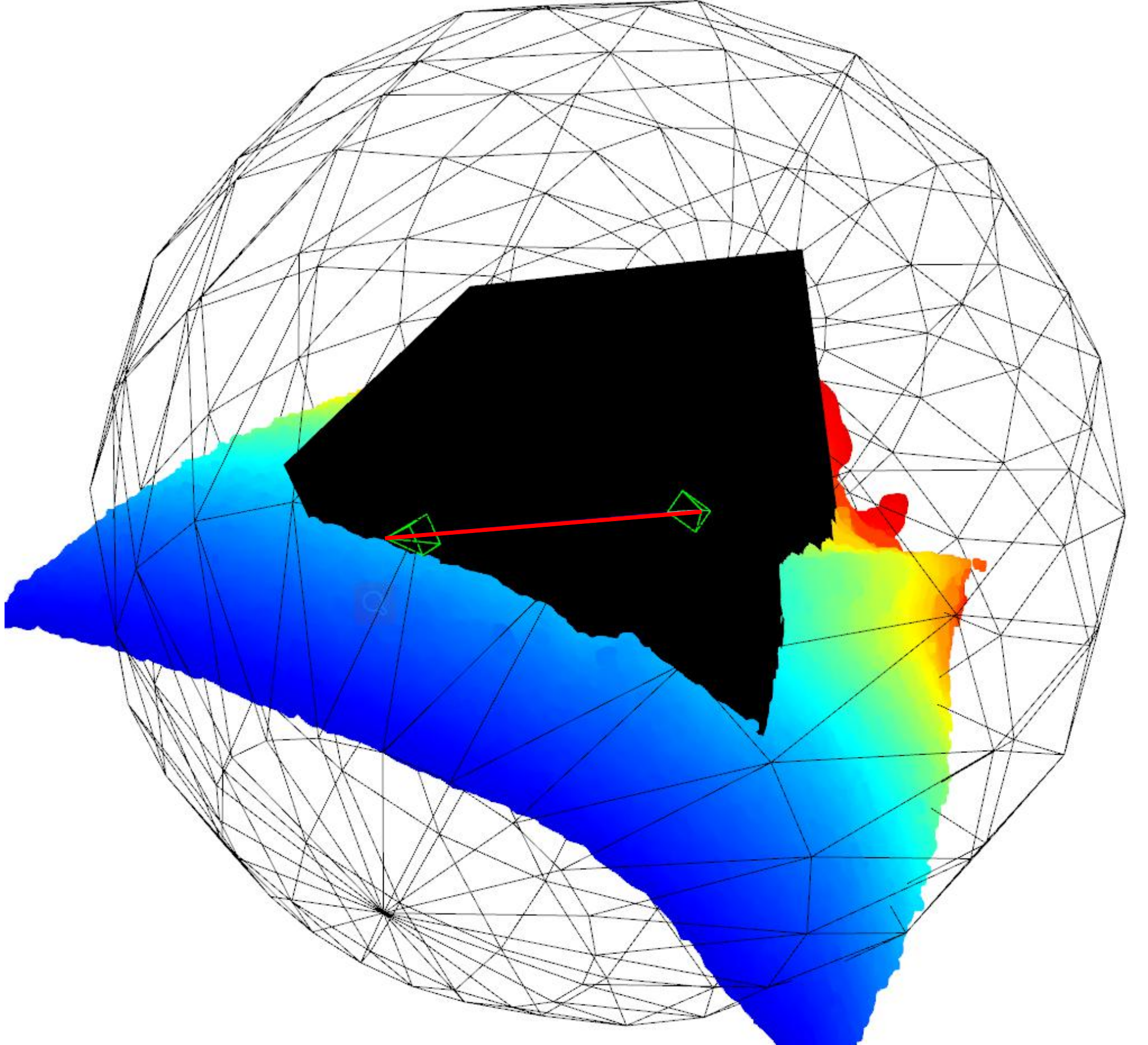}
    \caption{\textbf{The Visualization of Two View Initialization for Inside-forward Scenes.}}
    \label{fig:two_init_inside}
\end{figure}
\begin{figure}
    \centering
    \includegraphics[width=0.8\linewidth]{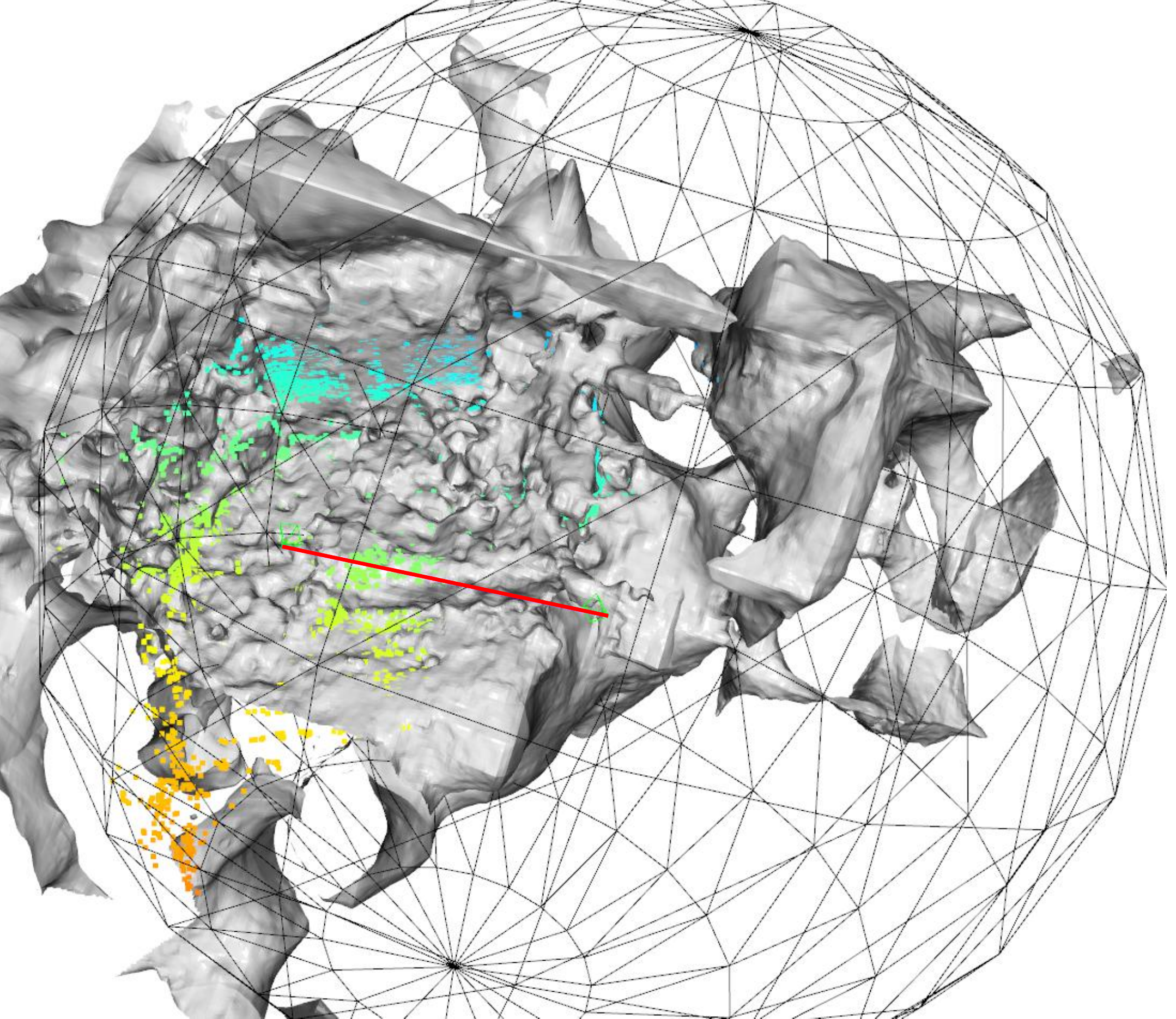}
    \caption{\textbf{The Visualization of Two View Initialization for Outside-forward Scenes.}}
    \label{fig:two_init_outside}
\end{figure}
\begin{figure}
    \centering
    \includegraphics[width=0.8\linewidth]{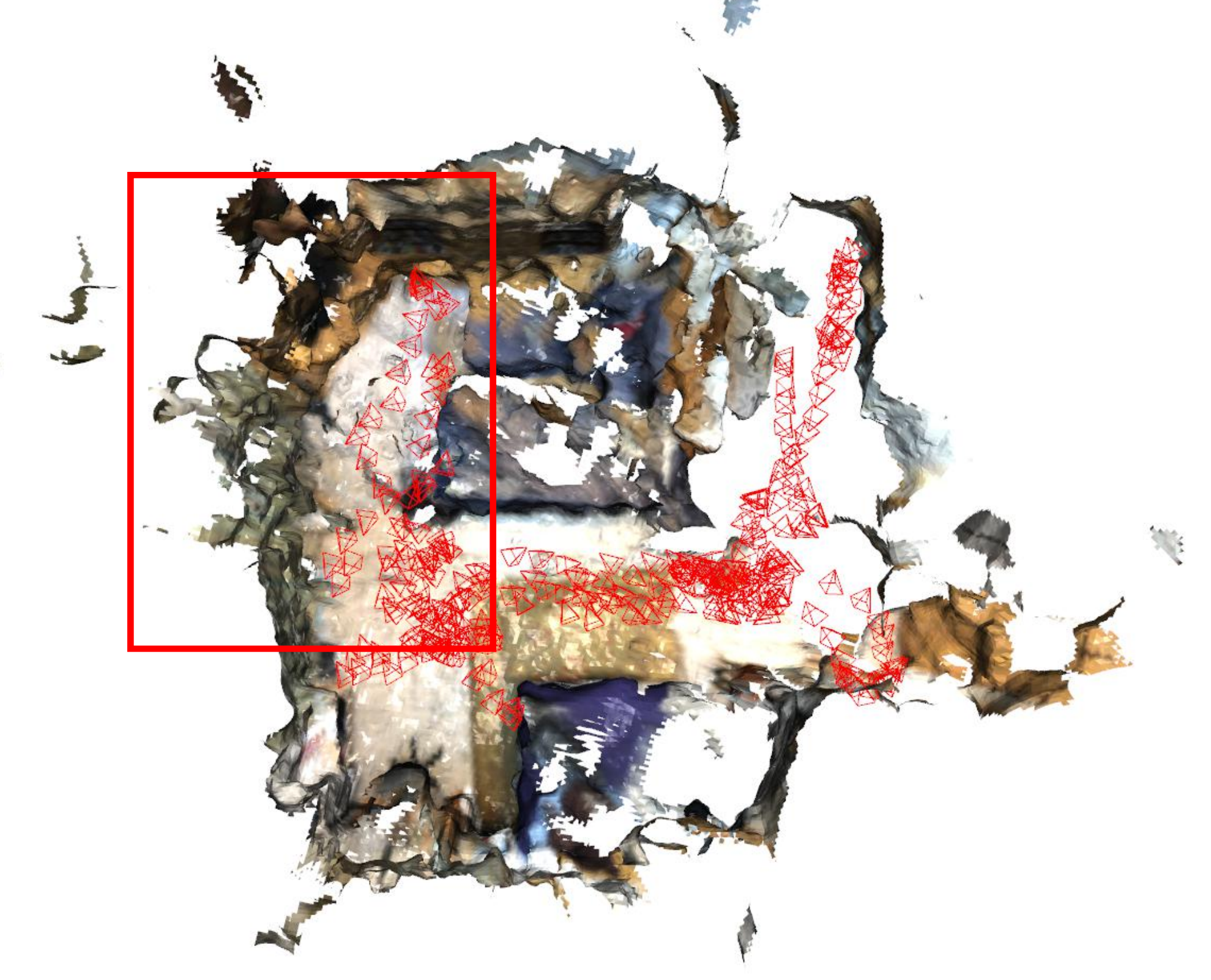}
    \caption{\textbf{Visualization of Mesh and Trajectory for Failure Cases.}This figure show the visualization of mesh and trajectory in scannet. As we can see the pose and geometry begin to be bad in the red box where the texture and matches are mostly less.}
    \label{fig:scan0000}
\end{figure}

Because the learning of neural implicit fields was originally designed for bounded scenes, we have to carefully design the two-view initialization for our Level-S$^2$fM and ensure the incremental reconstruction process is within the bound. In our study, we mainly focus on two representative types of scenes. 
The first type is the inside-forward scene, where the cameras are surrounded by the target object and inside forwarding (Specifically seen in Figure.~\ref{fig:two_init_inside}).
The second type is the outside-forward scenes. In initialization details for inside-forward scenes, we put the first camera on a sphere with $r=3$, and orient the camera toward the origin of the coordinate. The bound of the features grid is set to $[4,4,4]$. The initialized pose for the first camera is calculated by the following:

\begin{equation}
\begin{split}
\mathbf{t}^{\text{init}}_{\text{w2c}}&=\mathbf{R}^{\text{init}}_{\text{w2c}}\mathbf{t}^{\text{init}}_{\text{c2w}},\\
\mathbf{t}^{\text{init}}_{\text{c2w}}&=
\begin{bmatrix}
-r\cos\theta_y\cos\theta_z\\-r\cos\theta_y\sin\theta_z\\-r\sin\theta_y\\
\end{bmatrix},\\
\mathbf{R}^{\text{init}}_{\text{w2c}}&=\mathbf{R}_x(\theta_x)^{-1}\mathbf{{R}_y(\theta_y)}^{-1}{\mathbf{R}_z(\theta_z)}^{-1},\\
\end{split}
\label{equ:init_pose}
\end{equation}
where the $\theta_x=0,\theta_y=-\frac{1}{4}\pi,\theta_z=\frac{1}{4}\pi$. After that, the pose of the second camera is then initialized with the calculated relative pose by five points algorithm~\cite{5points}. Meanwhile, the length of the translation of the relative pose would be the hyper-parameter for different scenes. As shown in Figure.~\ref{fig:two_init_inside}, the red line is the baseline of the two-view camera, and the length of the baseline is empirically set. Moreover, for the outside-forward scene, the $\theta_x$ is set to $\frac{1}{2}\pi$ to make the orientation of the camera outside. It needs to be noticed that the specific parameters of two-view initialization may be different for different scenes, which can be referred into our codes and configuration after it is released. 
\section{Sphere Tracing and Depth Consistency}
\begin{figure}
    \centering
    \includegraphics[width=0.8\linewidth]{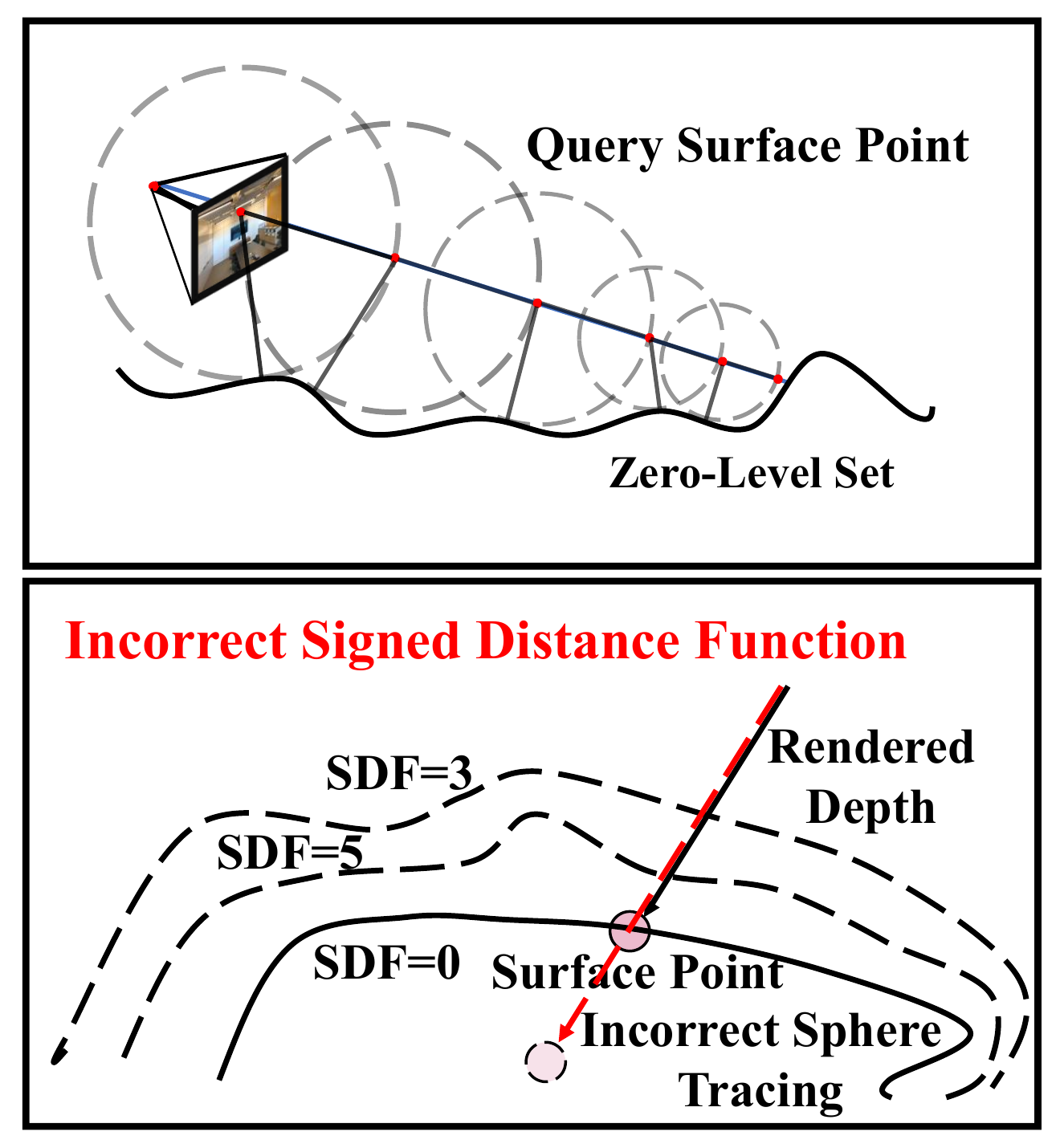}
    \caption{\textbf{Sphere Tracing and Depth Consistency.}}
    \label{fig:spheretracing}
\end{figure}

As mentioned in our paper, the key component for attaining the 3D points from the 2D keypoints is sphere tracing. As shown in Figure.~\ref{fig:spheretracing}, sphere tracing algorithm~\cite{dist_sphere} leverages the basic property of the signed distance function where the queried SDF value at each position is the closest distance from the point to the zero-level set of the surface. The depth of the queried 2d points can be calculated as follows:
\begin{equation}
\begin{split}
&\hat{t}=t_0+\sum_{j}^{max}sdf(\mathbf{X}_j),  \\
& X_{j+1}=X_{j}+sdf(X_{j})d,
\end{split}
\label{equ:sphere_tracing}
\end{equation}
where the start point $X_0=o+t_{0}d$. By the sphere tracing algorithm, we can efficiently get the 3d points from 2d, and it can be a natural constraint for the learning of SDF. But sometimes, sphere tracing can not correctly trace the surface when the zero-level set is correct while non-zero-level sets are wrong. Therefore, it can not ensure the multi-view consistency of the sphere tracing algorithm (seen in the second row of Figure.~\ref{fig:spheretracing}). To overcome that, we use the depth calculated by volumetric rendering as a constraint to keep the consistency between these two sampling strategies as mentioned in our paper.

\begin{figure*}
    \centering
    \includegraphics[width=0.96\linewidth]{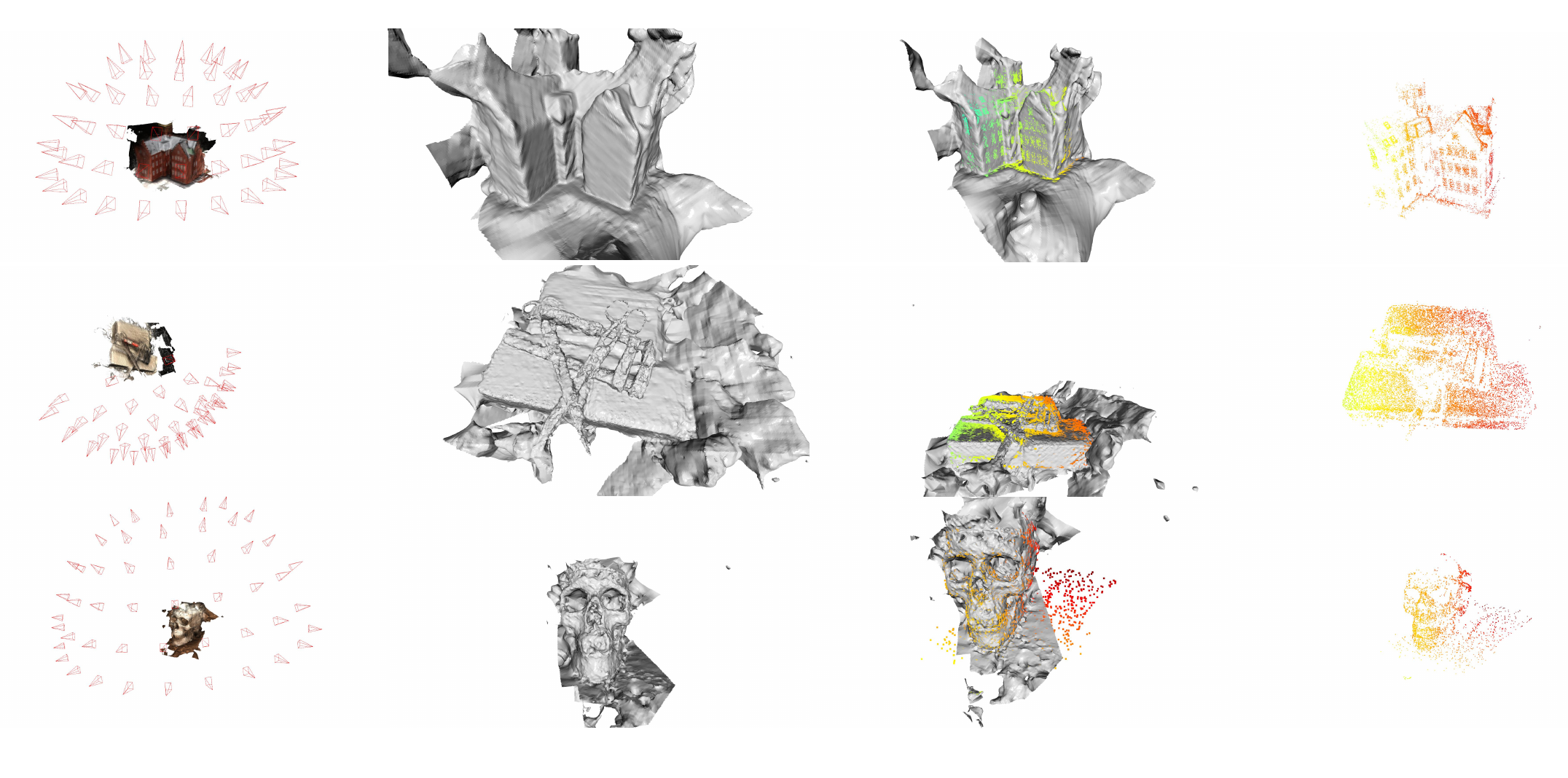}
    \caption{\textbf{Qualitative Results for Reconstruction and Pose Estimation.} The first column is the refused mesh and the visualization of camera poses. While, the second and third are the mesh and mesh shown with the points respectively. We can observe that our reconstructed points sticking on the surface of mesh. In the last column is the point cloud reconstructed.}
    \label{fig:supp_vis1}
\end{figure*}
\begin{figure}
    \centering
    \includegraphics[width=0.96\linewidth]{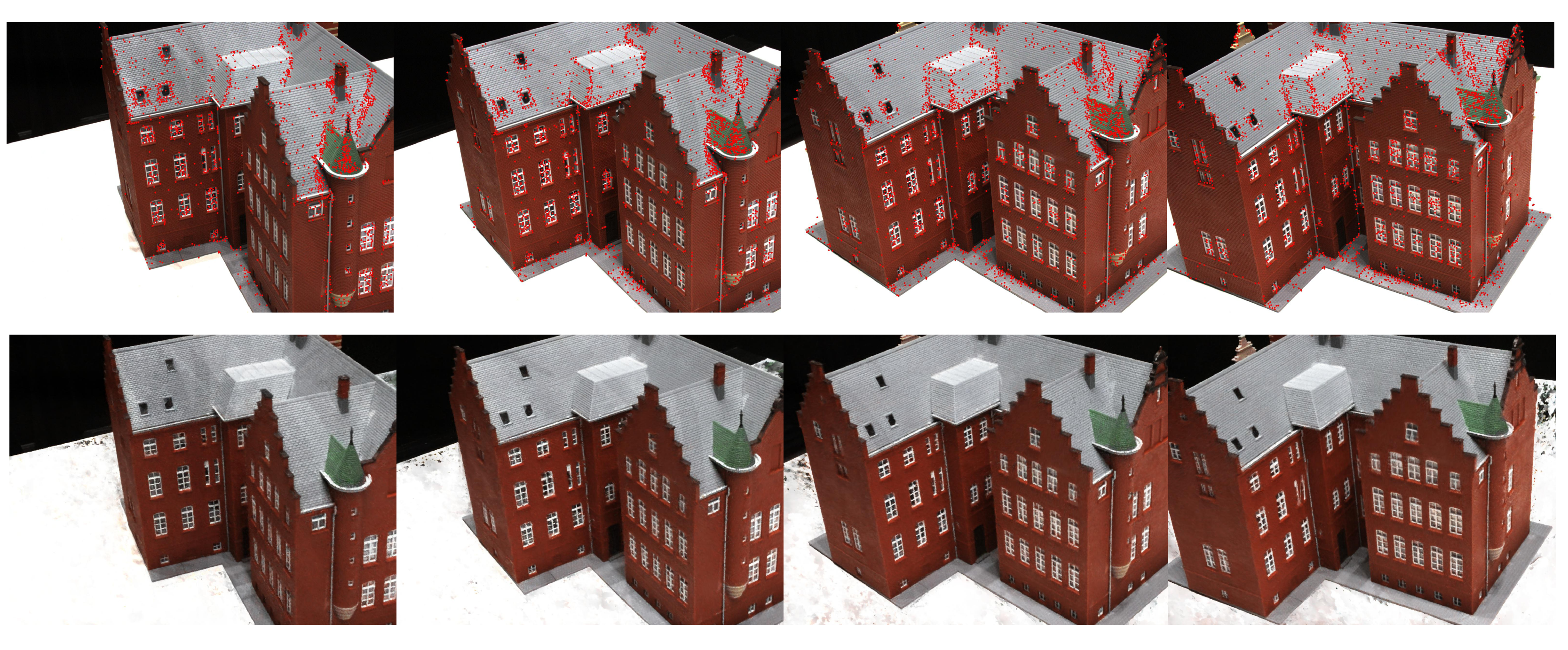}\\
    \includegraphics[width=0.96\linewidth]{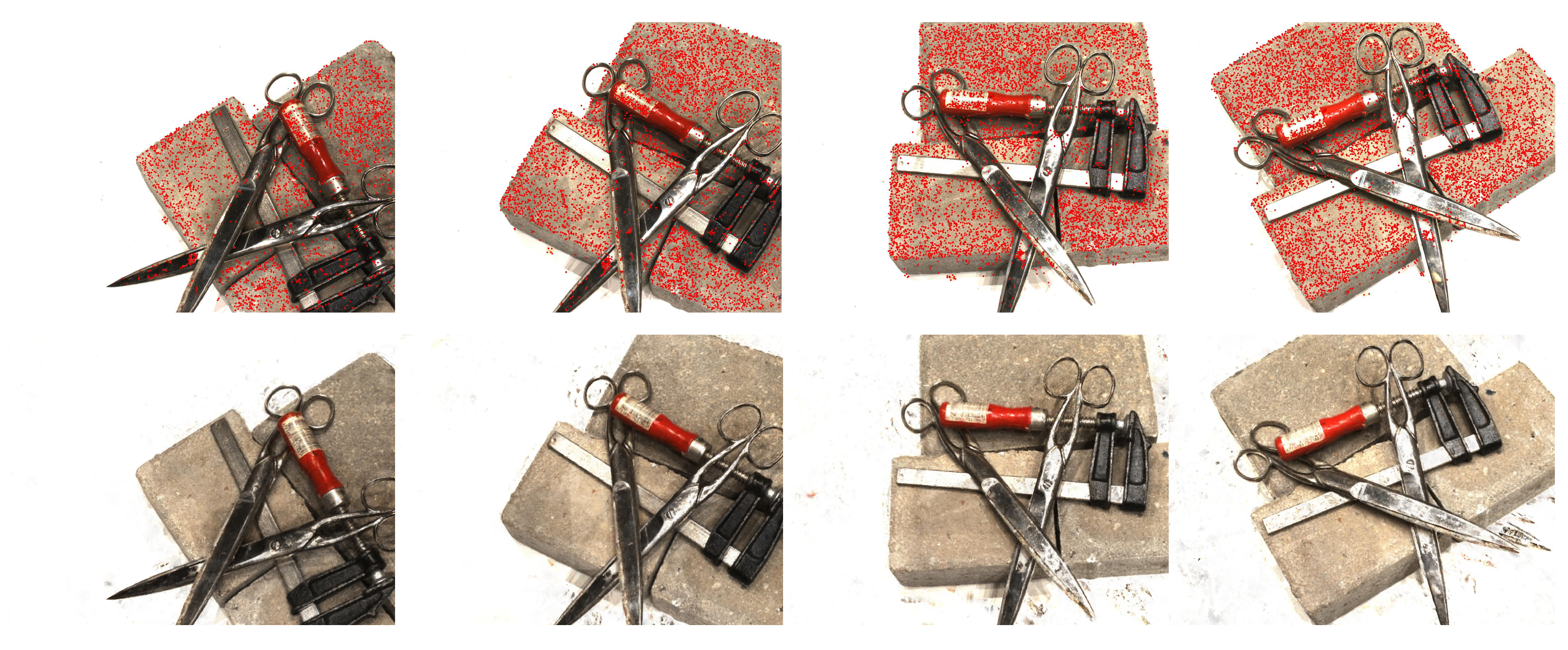}\\
    \includegraphics[width=0.96\linewidth]{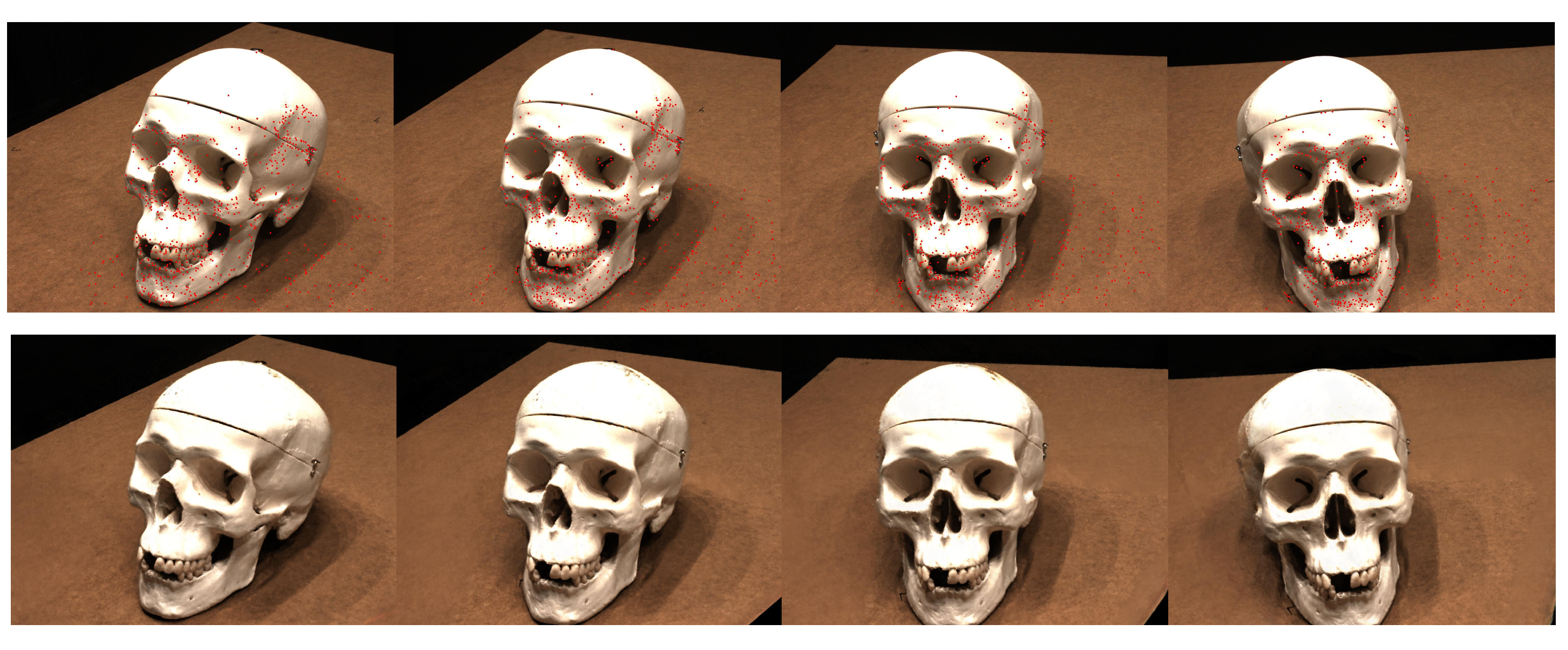}\\
    \caption{\textbf{Qualitative Results for Rendering.}This figure show the visualization of three scenes. At each group, the first row is the ground truth rgb images and their corresponding 3d observation projected on them. And the second row is the rendered images.}
    \label{fig:supp_vis}
\end{figure}
\begin{figure}
    \centering
    \includegraphics[width=0.96\linewidth]{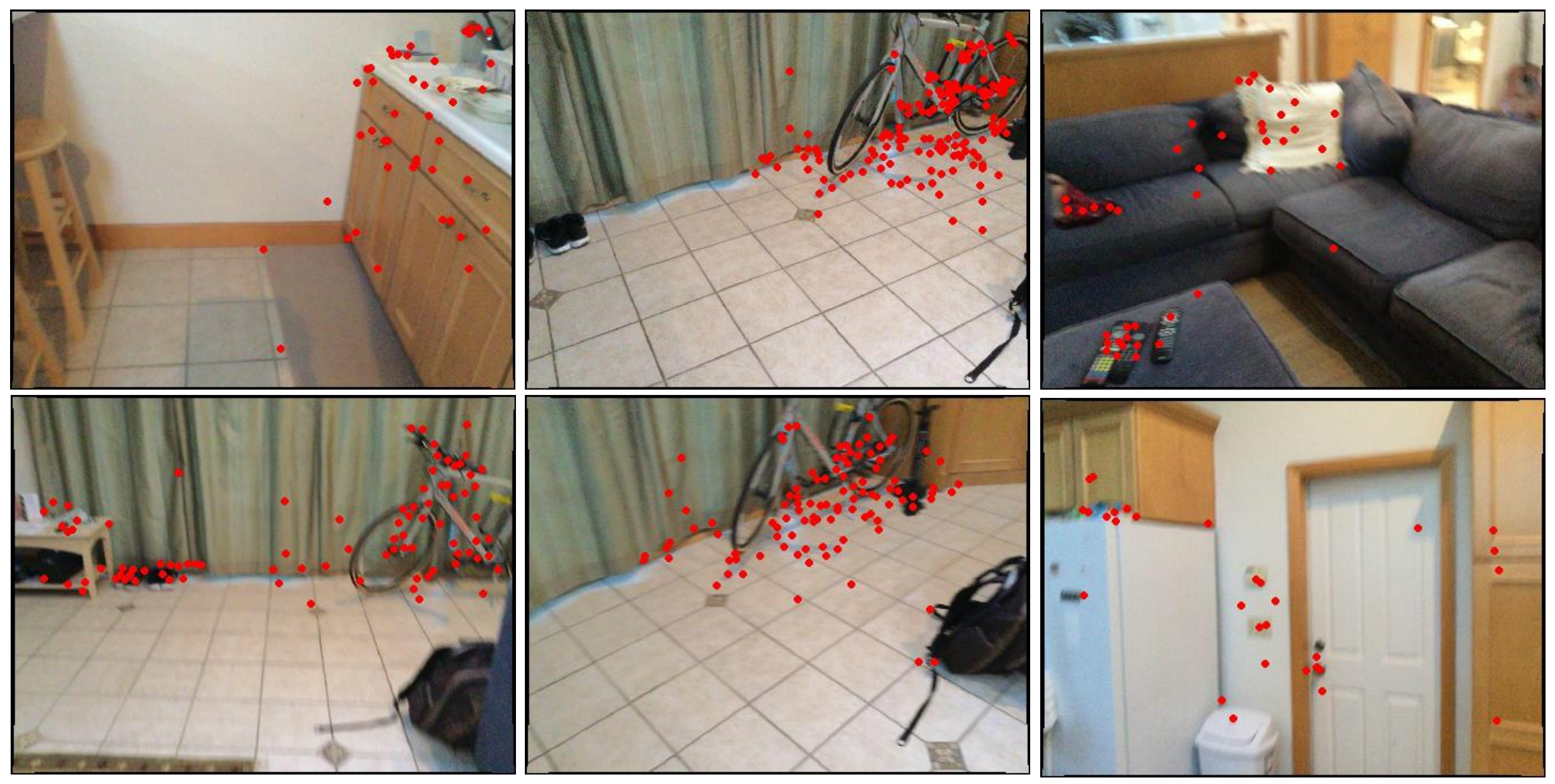}\\
    \includegraphics[width=0.96\linewidth]{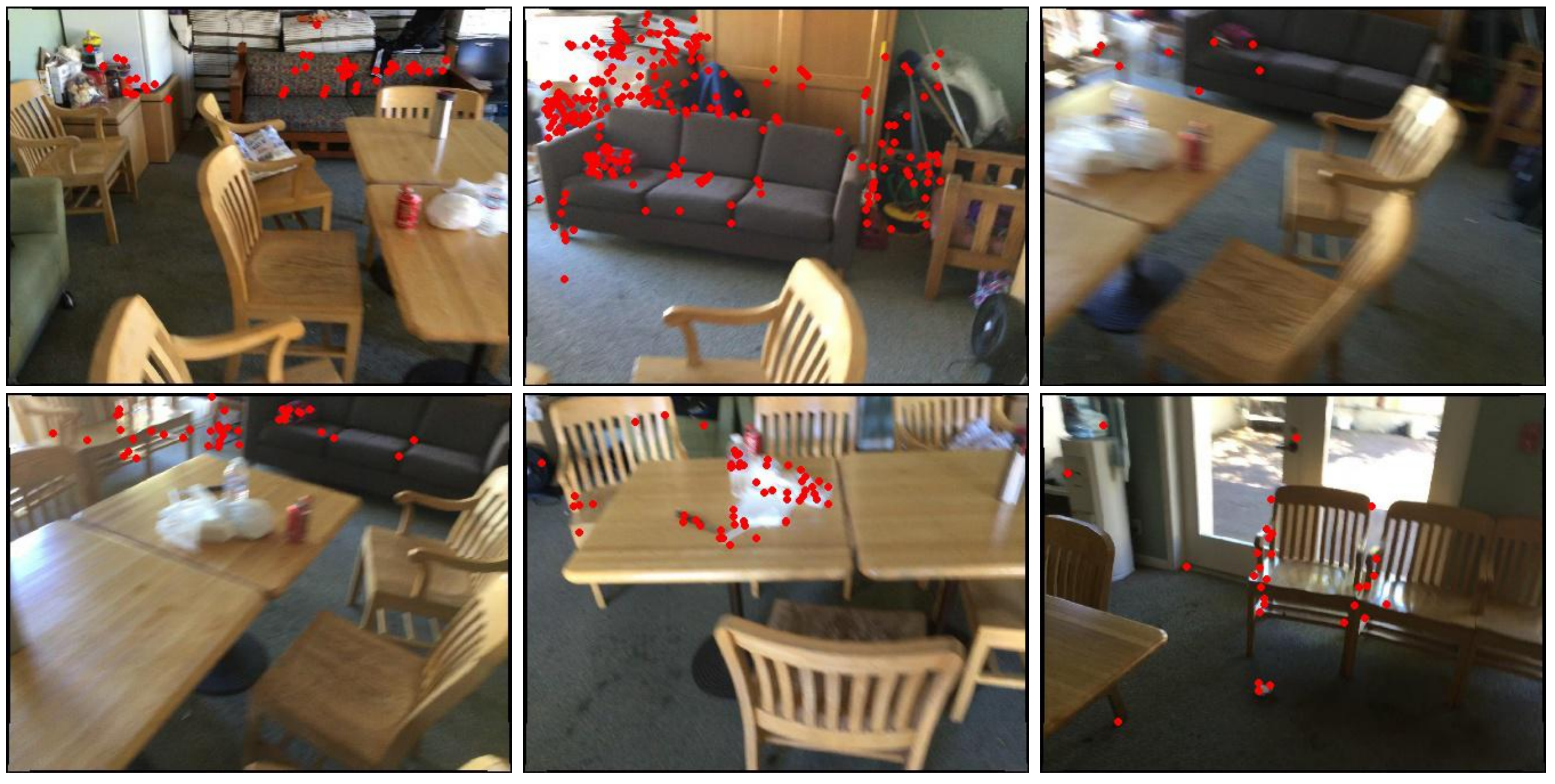}\\
    \caption{\textbf{Visualization for Images with Its 3D observations for PnP in Scannet.}This figure show that because of the blurry problem and textureless region, the 3D-2D correspondences are badly distributed and limited number.}
    \label{fig:supp_fail}
\end{figure}
\section{Sequence Order for Incremental Reconstruction}
For incremental SfM, the sequence order for incremental reconstruction is a relatively important component of the final result. But this paper concentrates our attention into renew the SfM on the neural level sets, which show its promising future to make breakthroughs. To avoid being exhausted to be stuck in the discussion of various tricks and strategies, we simply implement the next best view selection according to the number of 3D-2D pairs in PnP which may have a better alternative discussed in ~\cite{COLMAP}. In order completely discuss our framework, we also report the simple ablation result for the sequence order in \ours, which can be seen in Table.~\ref{table:ord}. We conducted the ablation study for sequence order in DTU~\cite{DTU}. We report two different sequence orders by randomly selecting the start of two frames for the two-view initialization. We can see that different sequence orders will cause different results. 
Despite of this, we would like to emphasize again that because of the complexity of Structure-from-Motion, the problem of the next-best view is not the core of our current study, which will be left in our future work. 

\begin{table}
\centering
\begin{tabular}{cc|c}
\toprule
\multicolumn{2}{c}{Scene}&\multicolumn{1}{c}{DTU~\cite{DTU}}\\
\hline
\multirow{2}{*}{\makecell{\textbf{Order 1st}}}&{\small rot\degree $\downarrow$}&{0.74}
\\&{\small trans. (mm) $\downarrow$}&{5.81}\\
\hline
\multirow{2}{*}{\makecell{\textbf{Order 2nd}}}&{\small rot\degree $\downarrow$}&{0.30}
\\&{\small trans. (mm) $\downarrow$}&{2.04}\\
\bottomrule 
\end{tabular}
\caption{{\bf Ablation for Sequence Order.} 
}
\vspace{-3mm}
\label{table:ord}
\end{table}

\section{More Qualitative Results on Individual Datasets}
We also report more qualitative results for our experiments. In Figure.~\ref{fig:supp_vis}, there are the rendered image results from our radiance field. While, in Figure.~\ref{fig:supp_vis1}, the estimated pose, reconstructed 3d points, and the mesh are visualized.

\section{Discussion on Failure Cases}
In our paper, we take the indoor datasets, ScanNet~\cite{scannet} to discuss our failure cases, where there are a lot of textureless regions. As shown in Figure.~\ref{fig:supp_fail}, because of the textureless areas or the blurry problem of the captured images, the 3D-2D correspondences are badly distributed and limited in number. Therefore, the registration of these images is hard to solve and results in bad initialization for the pose estimation. As seen in Figure.~\ref{fig:scan0000}, the trajectory of cameras becomes unsatisfactory due to the textureless wall. Meanwhile, due to the incremental reconstruction fashion, the subsequent pose is based on the former, so, the incorrect pose estimation results by the textureless region will influence the whole process. To alleviate the problem, the recent robust deep learning-based 2d matches method may play a core role~\cite{superpoint,superglue}, which will be our future works to explore the solution to this problem.

Meanwhile, for those NeRF~\cite{nerf} based SLAMs Framework~\cite{niceslam,imap}, they usually need the depth as an extra input. With the assistance of the depth map, the coarse pose is easily attained by aligning the depth of two consecutive frames, and they are not easily influenced by the issue of textureless. Therefore, in our paper, we did not compare our method with the depth-aware SLAM methods. Besides, we find that the optimizer for the networks is another core for getting stable and accurate pose results. The optimizer, Adam~\cite{adam}, used in our paper is may not the best choice for our problem as it is not easy to judge whether the optimization is converged. Therefore, in our future work, we are going to explore using the second-order optimizer like Levenberg-Marquardt (LM)~\cite{LMA} or Gauss-Newton (GN)~\cite{GNA} algorithm to solve our problem.  

\newpage
{\small
\bibliographystyle{ieee_fullname}
\bibliography{egbib}

\begin{thebibliography}{10}\itemsep=-1pt

\bibitem{barron2021mip}
Jonathan~T. Barron, Ben Mildenhall, Matthew Tancik, Peter Hedman, Ricardo
  Martin{-}Brualla, and Pratul~P. Srinivasan.
\newblock Mip-nerf: {A} multiscale representation for anti-aliasing neural
  radiance fields.
\newblock In {\em IEEE International Conference on Computer Vision (ICCV)},
  pages 5835--5844, 2021.

\bibitem{volumetricba}
Ronald Clark.
\newblock Volumetric bundle adjustment for online photorealistic scene capture.
\newblock In {\em IEEE Conference on Computer Vision and Pattern Recognition
  (CVPR)}, pages 6114--6122, 2022.

\bibitem{LSNet}
Ronald Clark, Michael Bloesch, Jan Czarnowski, Stefan Leutenegger, and
  Andrew~J. Davison.
\newblock Learning to solve nonlinear least squares for monocular stereo.
\newblock In {\em European Conference on Computer Vision (ECCV)}, volume 11212,
  pages 291--306, 2018.

\bibitem{globalsfm0}
David~J. Crandall, Andrew Owens, Noah Snavely, and Daniel~P. Huttenlocher.
\newblock Sfm with mrfs: Discrete-continuous optimization for large-scale
  structure from motion.
\newblock {\em IEEE Trans. on Pattern Analysis and Machine Intelligence
  (PAMI)}, 35(12):2841--2853, 2013.

\bibitem{cui2017hsfm}
Hainan Cui, Xiang Gao, Shuhan Shen, and Zhanyi Hu.
\newblock Hsfm: Hybrid structure-from-motion.
\newblock In {\em IEEE Conference on Computer Vision and Pattern Recognition
  (CVPR)}, pages 2393--2402, 2017.

\bibitem{scannet}
Angela Dai, Angel~X. Chang, Manolis Savva, Maciej Halber, Thomas~A. Funkhouser,
  and Matthias Nie{\ss}ner.
\newblock Scannet: Richly-annotated 3d reconstructions of indoor scenes.
\newblock In {\em IEEE Conference on Computer Vision and Pattern Recognition
  (CVPR)}, pages 2432--2443, 2017.

\bibitem{superpoint}
Daniel DeTone, Tomasz Malisiewicz, and Andrew Rabinovich.
\newblock Superpoint: Self-supervised interest point detection and description.
\newblock In {\em IEEE Conference on Computer Vision and Pattern Recognition
  Workshops (CVPRW)}, pages 224--236, 2018.

\bibitem{Ransac}
Martin~A Fischler and Robert~C Bolles.
\newblock Random sample consensus: a paradigm for model fitting with
  applications to image analysis and automated cartography.
\newblock {\em Communications of the ACM}, 24(6):381--395, 1981.

\bibitem{monodepth}
Cl{\'{e}}ment Godard, Oisin~Mac Aodha, Michael Firman, and Gabriel~J. Brostow.
\newblock Digging into self-supervised monocular depth estimation.
\newblock In {\em IEEE International Conference on Computer Vision (ICCV)},
  pages 3827--3837, 2019.

\bibitem{1996sphere}
John~C. Hart.
\newblock Sphere tracing: a geometric method for the antialiased ray tracing of
  implicit surfaces.
\newblock {\em The Visual Computer}, 12(10):527--545, 1996.

\bibitem{GNA}
Herman~O Hartley.
\newblock The modified gauss-newton method for the fitting of non-linear
  regression functions by least squares.
\newblock {\em Technometrics}, 3(2):269--280, 1961.

\bibitem{mvg}
Richard Hartley and Andrew Zisserman.
\newblock {\em Multiple view geometry in computer vision}.
\newblock Cambridge University Press, 2003.

\bibitem{triangulation}
Richard~I Hartley and Peter Sturm.
\newblock Triangulation.
\newblock {\em Computer Vision and Image Understanding (CVIU)}, 68(2):146--157,
  1997.

\bibitem{HrubyDLP22}
Petr Hruby, Timothy Duff, Anton Leykin, and Tom{\'{a}}s Pajdla.
\newblock Learning to solve hard minimal problems.
\newblock In {\em IEEE Conference on Computer Vision and Pattern Recognition
  (CVPR)}, pages 5522--5532, 2022.

\bibitem{DTU}
Rasmus~Ramsb{\o}l Jensen, Anders~Lindbjerg Dahl, George Vogiatzis, Engin Tola,
  and Henrik Aan{\ae}s.
\newblock Large scale multi-view stereopsis evaluation.
\newblock In {\em IEEE Conference on Computer Vision and Pattern Recognition
  (CVPR)}, pages 406--413, 2014.

\bibitem{kazhdan2006poisson}
Michael~M. Kazhdan, Matthew Bolitho, and Hugues Hoppe.
\newblock Poisson surface reconstruction.
\newblock In {\em Symposium on Geometry Processing}, volume 256, pages 61--70,
  2006.

\bibitem{adam}
Diederik~P Kingma and Jimmy Ba.
\newblock Adam: {A Method for Stochastic Optimization}.
\newblock In {\em International Conference on Learning Representations (ICLR)},
  2015.

\bibitem{epnp}
Vincent Lepetit, Francesc Moreno{-}Noguer, and Pascal Fua.
\newblock Ep\emph{n}p: An accurate \emph{O}(\emph{n}) solution to the
  p\emph{n}p problem.
\newblock {\em International Journal of Computer Vision (IJCV)},
  81(2):155--166, 2009.

\bibitem{barf}
Chen{-}Hsuan Lin, Wei{-}Chiu Ma, Antonio Torralba, and Simon Lucey.
\newblock {BARF:} bundle-adjusting neural radiance fields.
\newblock In {\em IEEE International Conference on Computer Vision (ICCV)},
  pages 5721--5731, 2021.

\bibitem{dist_sphere}
Shaohui Liu, Yinda Zhang, Songyou Peng, Boxin Shi, Marc Pollefeys, and Zhaopeng
  Cui.
\newblock {DIST:} rendering deep implicit signed distance function with
  differentiable sphere tracing.
\newblock In {\em IEEE Conference on Computer Vision and Pattern Recognition
  (CVPR)}, pages 2016--2025, 2020.

\bibitem{MC}
William~E. Lorensen and Harvey~E. Cline.
\newblock Marching cubes: {A} high resolution 3d surface construction
  algorithm.
\newblock In {\em Proceedings of {SIGGRAPH}}, pages 163--169, 1987.

\bibitem{SIFT}
David~G Lowe.
\newblock Distinctive image features from scale-invariant keypoints.
\newblock {\em International Journal of Computer Vision (IJCV)}, 60(2):91--110,
  2004.

\bibitem{sfmlearner}
Reza Mahjourian, Martin Wicke, and Anelia Angelova.
\newblock {Unsupervised Learning of Depth and Ego-Motion From Monocular Video
  Using 3D Geometric Constraints}.
\newblock In {\em IEEE Conference on Computer Vision and Pattern Recognition
  (CVPR)}, pages 5667--5675, 2018.

\bibitem{levelset-evolution}
Ishit Mehta, Manmohan Chandraker, and Ravi Ramamoorthi.
\newblock A level set theory for neural implicit evolution under explicit
  flows.
\newblock In {\em European Conference on Computer Vision (ECCV)}, 2022.

\bibitem{nerf}
Ben Mildenhall, Pratul~P. Srinivasan, Matthew Tancik, Jonathan~T. Barron, Ravi
  Ramamoorthi, and Ren Ng.
\newblock Nerf: Representing scenes as neural radiance fields for view
  synthesis.
\newblock In {\em European Conference on Computer Vision (ECCV)}, volume 12346,
  pages 405--421, 2020.

\bibitem{LMA}
Jorge~J Mor{\'e}.
\newblock The levenberg-marquardt algorithm: implementation and theory.
\newblock In {\em Numerical analysis}, pages 105--116. 1978.

\bibitem{tiny-cuda-nn}
Thomas M\"uller.
\newblock {tiny-cuda-nn}, 4 2021.

\bibitem{instantngp}
Thomas M{\"{u}}ller, Alex Evans, Christoph Schied, and Alexander Keller.
\newblock Instant neural graphics primitives with a multiresolution hash
  encoding.
\newblock {\em ACM Trans. on Graphics (TOG)}, 41(4):102:1--102:15, 2022.

\bibitem{5points}
David Nist{\'{e}}r.
\newblock {An Efficient Solution to the Five-Point Relative Pose Problem}.
\newblock {\em IEEE Trans. on Pattern Analysis and Machine Intelligence
  (PAMI)}, 26(6):756--777, 2004.

\bibitem{unisurf}
Michael Oechsle, Songyou Peng, and Andreas Geiger.
\newblock {UNISURF:} unifying neural implicit surfaces and radiance fields for
  multi-view reconstruction.
\newblock In {\em IEEE International Conference on Computer Vision (ICCV)},
  pages 5569--5579, 2021.

\bibitem{deepsdf}
Jeong~Joon Park, Peter Florence, Julian Straub, Richard~A. Newcombe, and Steven
  Lovegrove.
\newblock Deepsdf: Learning continuous signed distance functions for shape
  representation.
\newblock In {\em IEEE Conference on Computer Vision and Pattern Recognition
  (CVPR)}, pages 165--174, 2019.

\bibitem{pytorch}
Adam Paszke, Sam Gross, Francisco Massa, Adam Lerer, James Bradbury, Gregory
  Chanan, Trevor Killeen, Zeming Lin, Natalia Gimelshein, Luca Antiga, Alban
  Desmaison, Andreas K{\"{o}}pf, Edward~Z. Yang, Zachary DeVito, Martin Raison,
  Alykhan Tejani, Sasank Chilamkurthy, Benoit Steiner, Lu Fang, Junjie Bai, and
  Soumith Chintala.
\newblock Pytorch: An imperative style, high-performance deep learning library.
\newblock In {\em Advances in Neural Information Processing Systems (NeurIPS)},
  pages 8024--8035, 2019.

\bibitem{superglue}
Paul{-}Edouard Sarlin, Daniel DeTone, Tomasz Malisiewicz, and Andrew
  Rabinovich.
\newblock Superglue: Learning feature matching with graph neural networks.
\newblock In {\em IEEE Conference on Computer Vision and Pattern Recognition
  (CVPR)}, pages 4937--4946, 2020.

\bibitem{COLMAP}
Johannes~L Sch\"{o}nberger and Jan-Michael Frahm.
\newblock {Structure-from-motion Revisited}.
\newblock In {\em IEEE Conference on Computer Vision and Pattern Recognition
  (CVPR)}, pages 4104--4113, 2016.

\bibitem{Patchmatch}
Johannes~L. Sch{\"{o}}nberger, Enliang Zheng, Jan{-}Michael Frahm, and Marc
  Pollefeys.
\newblock Pixelwise view selection for unstructured multi-view stereo.
\newblock In {\em European Conference on Computer Vision (ECCV)}, volume 9907,
  pages 501--518, 2016.

\bibitem{eth3d}
Thomas Sch{\"{o}}ps, Torsten Sattler, and Marc Pollefeys.
\newblock {BAD} {SLAM:} bundle adjusted direct {RGB-D} {SLAM}.
\newblock In {\em IEEE Conference on Computer Vision and Pattern Recognition
  (CVPR)}, pages 134--144, 2019.

\bibitem{scenegraph}
Noah Snavely, Steven~M. Seitz, and Richard Szeliski.
\newblock Skeletal graphs for efficient structure from motion.
\newblock In {\em IEEE Conference on Computer Vision and Pattern Recognition
  (CVPR)}, 2008.

\bibitem{rgbd_eval}
J{\"{u}}rgen Sturm, Nikolas Engelhard, Felix Endres, Wolfram Burgard, and
  Daniel Cremers.
\newblock A benchmark for the evaluation of {RGB-D} {SLAM} systems.
\newblock In {\em IEEE/RSJ International Conference on Intelligent Robots and
  Systems (IROS)}, pages 573--580, 2012.

\bibitem{imap}
Edgar Sucar, Shikun Liu, Joseph Ortiz, and Andrew~J. Davison.
\newblock imap: Implicit mapping and positioning in real-time.
\newblock In {\em IEEE International Conference on Computer Vision (ICCV)},
  pages 6209--6218, 2021.

\bibitem{BAnet}
Chengzhou Tang and Ping Tan.
\newblock {BA-Net: Dense Bundle Adjustment Networks}.
\newblock In {\em International Conference on Learning Representations (ICLR)},
  2019.

\bibitem{two_viewsfm}
Jianyuan Wang, Yiran Zhong, Yuchao Dai, Stan Birchfield, Kaihao Zhang, Nikolai
  Smolyanskiy, and Hongdong Li.
\newblock Deep two-view structure-from-motion revisited.
\newblock In {\em IEEE Conference on Computer Vision and Pattern Recognition
  (CVPR)}, pages 8953--8962, 2021.

\bibitem{TC-SfM}
Lei Wang, Linlin Ge, Shan Luo, Zihan Yan, Zhaopeng Cui, and Jieqing Feng.
\newblock Tc-sfm: Robust track-community-based structure-from-motion.
\newblock {\em CoRR}, abs/2206.05866, 2022.

\bibitem{neus}
Peng Wang, Lingjie Liu, Yuan Liu, Christian Theobalt, Taku Komura, and Wenping
  Wang.
\newblock Neus: Learning neural implicit surfaces by volume rendering for
  multi-view reconstruction.
\newblock In {\em Advances in Neural Information Processing Systems (NeurIPS)},
  pages 27171--27183, 2021.

\bibitem{HFS}
Yiqun Wang, Ivan Skorokhodov, and Peter Wonka.
\newblock Improved surface reconstruction using high-frequency details.
\newblock {\em CoRR}, abs/2206.07850, 2022.

\bibitem{nerf--}
Zirui Wang, Shangzhe Wu, Weidi Xie, Min Chen, and Victor~Adrian Prisacariu.
\newblock Nerf-: Neural radiance fields without known camera parameters.
\newblock {\em CoRR}, abs/2102.07064, 2021.

\bibitem{globalsfm1}
Kyle Wilson and Noah Snavely.
\newblock Robust global translations with 1dsfm.
\newblock In {\em European Conference on Computer Vision (ECCV)}, volume 8691,
  pages 61--75, 2014.

\bibitem{DeepMLE}
Yuxi Xiao, Li Li, Xiaodi Li, and Jian Yao.
\newblock Deepmle: {A} robust deep maximum likelihood estimator for two-view
  structure from motion.
\newblock In {\em IEEE/RSJ International Conference on Intelligent Robots and
  Systems (IROS)}, pages 10643--10650, 2022.

\bibitem{blendedmvs}
Yao Yao, Zixin Luo, Shiwei Li, Jingyang Zhang, Yufan Ren, Lei Zhou, Tian Fang,
  and Long Quan.
\newblock Blendedmvs: {A} large-scale dataset for generalized multi-view stereo
  networks.
\newblock In {\em IEEE Conference on Computer Vision and Pattern Recognition
  (CVPR)}, pages 1787--1796, 2020.

\bibitem{volsdf}
Lior Yariv, Jiatao Gu, Yoni Kasten, and Yaron Lipman.
\newblock Volume rendering of neural implicit surfaces.
\newblock In {\em Advances in Neural Information Processing Systems (NeurIPS)},
  pages 4805--4815, 2021.

\bibitem{niceslam}
Zihan Zhu, Songyou Peng, Viktor Larsson, Weiwei Xu, Hujun Bao, Zhaopeng Cui,
  Martin~R. Oswald, and Marc Pollefeys.
\newblock {NICE-SLAM:} neural implicit scalable encoding for {SLAM}.
\newblock In {\em IEEE Conference on Computer Vision and Pattern Recognition
  (CVPR)}, pages 12776--12786, 2022.

\end{thebibliography}
}

\end{document}